\documentclass{article}

    \PassOptionsToPackage{numbers, compress}{natbib}
\usepackage[main, final]{neurips_2026} % CAMERA-READY: final option de-anonymizes and sets the footer

\usepackage[utf8]{inputenc} % allow utf-8 input
\usepackage[T1]{fontenc}    % use 8-bit T1 fonts
\usepackage{hyperref}       % hyperlinks
\usepackage{url}            % simple URL typesetting
\usepackage{booktabs}       % professional-quality tables
\usepackage{amsfonts}       % blackboard math symbols
\usepackage{nicefrac}       % compact symbols for 1/2, etc.
\usepackage{microtype}      % microtypography
\usepackage{xcolor}         % colors
\usepackage{xspace}
\usepackage{natbib}
\usepackage{graphicx}
\usepackage{amsmath}
\usepackage{amsthm}
\usepackage{hyperref}
\usepackage{url}
\usepackage{xspace}
\usepackage{graphicx}
\usepackage{hyperref}
\usepackage{url}
\usepackage{multirow}
\usepackage{graphicx}
\usepackage{amsmath}
\usepackage{amsthm}
\usepackage{booktabs}
\usepackage{subcaption}
\usepackage{algorithm}
\usepackage{algorithmic}
\usepackage{amsfonts,amssymb}
\usepackage{hhline}
\usepackage{diagbox}
\usepackage{xcolor}
\usepackage{adjustbox}
\usepackage{multirow}
\usepackage{graphicx}
\usepackage{subcaption}
\usepackage{tabularx}
\usepackage{booktabs}
\usepackage[capitalize]{cleveref}
\usepackage{bigstrut}
\usepackage{rotating}
\usepackage{makecell}
\usepackage{xurl}
\usepackage{wrapfig}

\newif\ifshowchanges
\showchangestrue
\title{Steering Vectors as a Training Signal in LLM Post-Training}

\author{%
  Tiejin Chen$^{1}$ \quad
  Maunil Vyas$^{2}$\thanks{This work was conducted independently by the author and was not undertaken as part of their role at Amazon.} \quad
  Huaiyuan Yao$^{1}$ \quad
  Hua Wei$^{1}$ \\
  $^{1}$Arizona State University \quad $^{2}$Amazon \\
}

\begin{document}

\maketitle

\begin{abstract}
Steering vectors (SVs) are residual-stream directions that shift the behavior of LLMs when added to hidden states at inference time. While prior work has mainly used SVs as post-hoc controls, we ask whether the same directions can also guide parameter updates during post-training. We study this question across supervised learning, including SFT and distillation, and reinforcement learning with verifiable rewards (RLVR). Our framework considers two intervention regimes: a hidden-state hook during teacher-forced supervised training, and an advantage-repair mechanism on wrong rollouts in GRPO. For each regime, we analyze three design axes, which are the steering direction, its placement, and the operation used to combine it with training. In supervised settings, SV injection improves over plain SFT and distillation, with gains scaling with the off-policy gap and reaching up to a $12\%$ improvement when distilling from thinking-mode teachers. These gains hold across three random seeds. In RL settings, applying a correctness-contrast direction as advantage repair on wrong rollouts, including otherwise zero-gradient all-wrong groups, recovers a learning signal that transfers to held-out math benchmarks, and acts as a composable auxiliary signal that adds to existing zero-variance remedies such as entropy-aware advantage shaping. We further identify three failure boundaries. In detail, code generation can break the direction, cross-family layer transfer can break the placement, and inference-time application on an SV-trained checkpoint can break the intervention regime. Together, these results establish steering vectors as a conditional but effective training signal for LLM post-training. Our code is available \href{https://github.com/tiejin98/sv_training}{here}.
\end{abstract}

\section{Introduction}
\label{sec:intro}
A steering vector (SV) is a direction in an LLM's residual stream that can shift the model's behavior when added to the hidden state at inference time~\citep{turner2023activation, panickssery2024caa, zou2023representation}. Prior work has mainly used steering vectors as post-hoc controls for frozen models: the model parameters remain unchanged, and the vector is injected only during generation. This raises a natural question: \emph{if a steering vector captures a reusable behavioral direction, can it also be used to guide parameter updates during post-training?}

This question is important because post-training often relies on fixed reference data or sparse reward signals. In supervised fine-tuning and distillation~\citep{ouyang2022instructgpt, hinton2015distillation}, the model is optimized on reference trajectories rather than the trajectories induced by its own current policy, which creates a mismatch between training-time and inference-time behavior~\citep{ranzato2015sequence}. In reinforcement learning with verifiable rewards, the reward signal may be sparse, especially when all sampled responses in a group are incorrect. Steering vectors offer a representation-level intervention: instead of changing only the final output distribution at inference time, they can modify the hidden states used during training and thereby influence the gradients used to update the model. This turns steering from a temporary control mechanism into a possible training signal.

In this paper, we study this possibility systematically. We investigate how steering vectors can be integrated into both supervised post-training (SFT and knowledge distillation) and reinforcement learning with verifiable rewards (RLVR)~\citep{deepseek2025r1, shao2024deepseekmath}. We organize the study around two intervention regimes that correspond to where the SV enters the optimization pipeline, namely a \emph{hidden-state hook} during supervised post-training and an \emph{advantage repair} on wrong rollouts during GRPO. Within each regime we describe the design along three axes, namely the \emph{direction} itself, its \emph{placement}, and the \emph{operation} that combines it with the existing computation. Our findings reveal a structured, conditional picture. On the direction choice, per-prompt vectors are competitive with or outperform global means in supervised post-training, and the gain scales with the size of the off-policy gap, reaching up to a $12\%$ improvement when distilling from thinking-mode teachers. In the RL regime, the correctness-contrast direction applied as advantage repair on wrong rollouts, including otherwise zero-gradient all-wrong groups, recovers a learning signal that transfers cleanly to held-out math benchmarks. Overall, our contributions are as follows:

\noindent $\bullet$ We frame SV injection as a training signal across two intervention regimes (a hidden-state hook for supervised post-training and an advantage repair on wrong GRPO rollouts), and describe each regime along three axes, namely the \emph{direction} itself, its \emph{placement}, and the \emph{operation} that combines it with the existing computation. To our knowledge, this is the first systematic study of SVs as a persistent training signal rather than only an inference-time control.

\noindent $\bullet$ Across multiple students, datasets, and teachers, we identify configurations that reliably help in both regimes. In the supervised regime, we find that adding SVs during teacher-forced training generally improves over Plain SFT and distillation. In the RL regime, an advantage repair on wrong GRPO rollouts using a correctness-contrast direction (SV-Rescue) recovers a useful learning signal as a composable auxiliary signal that stacks with entropy-aware advantage shaping~\citep{lai2025noprompt}.

\noindent $\bullet$ We map the corresponding failure boundaries. Code generation breaks the \emph{direction}, where the reference direction conflicts with the test-passing direction; cross-family layer choice breaks the \emph{placement}; and inference-time application on an SV-trained checkpoint breaks the \emph{regime}, removing rather than amplifying the gain.

\section{Related Work}
\label{sec:related}

\paragraph{Steering Vectors.}
Steering vectors are directions in the residual stream of an LLM that, when added or otherwise mixed into the hidden state, predictably shift high-level model behavior without changing parameters~\citep{turner2023activation, li2023inference, panickssery2024caa, zou2023representation}. Related formulations include function vectors~\citep{todd2024fv} and weight-space task arithmetic~\citep{ilharco2023taskvec}, and a recent line of work refines their reliability, scaling, and geometry~\citep{imbrasaite2025reliability, zhao2025saersv, valois2025conceptors, tang2025ttsv, li2025textual, vyas2025thinking, subramani2025feedback, shoemake1985slerp, goddard2024mergekit, luo2026spherical}. All of these approaches use the steering vector as an inference-time control signal on a frozen model, and none tests its influence during the training stage. Our work fills this gap.

\paragraph{Post-training of LLMs.}
Modern LLM post-training starts from supervised fine-tuning (SFT) on reference data~\citep{ouyang2022instructgpt, wang2023selfinstruct,taori2023alpaca}, often with parameter-efficient adapters such as LoRA~\citep{hu2022lora}. SFT is also the main tool used to elicit chain-of-thought-style reasoning before reinforcement learning~\citep{wei2022cot, cobbe2021gsm8k, zelikman2022star, dong2023raft}. Knowledge distillation, which trains a student to imitate a teacher's output distribution~\citep{hinton2015distillation, gu2024minillm}, can be viewed as a special case of SFT on teacher-produced targets. A well-known limitation of teacher-forced objectives is off-policy gap~\citep{ranzato2015sequence}. DFT~\citep{wu2025dft} addresses this gap at the level of the objective by reweighting each token's loss with its detached probability, whereas our SV acts on the hidden representation. None of the previous SFT works considers using a steering vector during training. Beyond supervised methods, modern post-training pipelines increasingly adopt reinforcement learning with verifiable rewards (RLVR) to improve model capabilities~\citep{christiano2017deep, schulman2017ppo, lambert2024tulu3, deepseek2025r1, guo2025rlvrreason}. Prominent algorithms in this space include GRPO~\citep{shao2024deepseekmath} as a widely used critic-free variant. However, a known weakness of GRPO is the zero variance on the all-wrong group. Several recent methods address these groups through dynamic sampling, entropy-guided advantage shaping, or gold-rollout injection~\citep{yu2025dapo, lai2025noprompt, wu2025gcpo}. Our work fits into this family of rescue methods, but uniquely leverages SVs. Because it only shifts the advantage of wrong rollouts, it is complementary to these methods and can be stacked on top of them.

\section{Preliminaries}
\label{sec:background}

% This section reviews the two ingredients on which our method builds. 

% \paragraph{Notation.}
Let $\pi_\theta$ be an autoregressive language model with $L$ transformer blocks and hidden size $d$. For a prompt $x$ and a response $y = (y_1, \dots, y_T)$, let $h^{(\ell)}(x, y_{<t}) \in \mathbb{R}^d$ denote the residual-stream hidden state at the output of block $\ell \in \{1, \dots, L\}$ when processing the token at position $t$. We write $h^{(\ell)}_t$ when the context is clear, and $m \in \{0,1\}^T$ for the binary mask selecting answer tokens.

\subsection{Steering vectors}
\label{subsec:sv}

A broad empirical observation in the interpretability literature is that many high-level behaviors of LLMs, such as sentiment, truthfulness, refusal, and task identity, are encoded along approximately linear directions in the residual stream~\citep{turner2023activation, li2023inference, panickssery2024caa, zou2023representation}. This motivates the goal of finding, at each layer $\ell$, a single direction $v^{(\ell)} \in \mathbb{R}^d$ such that shifting the hidden state along $v^{(\ell)}$ predictably shifts the corresponding behavior in the model's output.

Given two matched response populations $\mathcal{D}_+$ and $\mathcal{D}_-$ that differ in the target behavior, the standard recipe~\citep{panickssery2024caa, zou2023representation} sets $v^{(\ell)}$ to the difference of the mean hidden states at layer $\ell$ over answer tokens:
\begin{equation}
v^{(\ell)} \;=\;
\mathbb{E}_{i \in \mathcal{D}_+}\!\!\left[
\bar h^{(\ell)}(x_i, y_i^{+}) \right]
\;-\;
\mathbb{E}_{i \in \mathcal{D}_-}\!\!\left[
\bar h^{(\ell)}(x_i, y_i^{-}) \right],
\label{eq:sv-generic}
\end{equation}
where $\bar h^{(\ell)}(x, y)$ denotes the mean hidden state at layer $\ell$ over the answer tokens of $y$. Appendix~\ref{sec:sv_appendix} gives the explicit pooling formula and the intuition.

\paragraph{Injection.}
There are multiple primitives for applying $v^{(\ell)}$ to the hidden state at a chosen layer $\ell^\star$. We introduce two here:  (1) \textit{Additive injection}~\citep{turner2023activation, panickssery2024caa} takes:
\begin{equation}
\tilde h^{(\ell^\star)}_t \;=\;
h^{(\ell^\star)}_t \;+\; \alpha \, v^{(\ell^\star)},
\qquad \alpha \in \mathbb{R},
\label{eq:additive-inject}
\end{equation}
but changes both the direction and the magnitude of the hidden state. (2) \textit{Spherical linear interpolation (Slerp)}~\citep{shoemake1985slerp, goddard2024mergekit, luo2026spherical} avoids changing magnitude by rotating the hidden state on the unit sphere toward a target direction along the shortest great-circle arc:
\begin{equation}
\mathrm{Slerp}(u, w; \alpha) \;=\;
\frac{\sin\!\big((1-\alpha)\omega\big)}{\sin \omega}\, u
\;+\;
\frac{\sin(\alpha \omega)}{\sin \omega}\, w,
\qquad \omega = \arccos(u^{\!\top} w).
\label{eq:slerp-def}
\end{equation}
More details can be found at Appendix~\ref{sec:sv_appendix}.

\subsection{Group Relative Policy Optimization (GRPO)}
\label{subsec:grpo}

For reasoning tasks with a verifier (e.g.\ final-answer matching on
GSM8K~\citep{cobbe2021gsm8k}), GRPO~\citep{shao2024deepseekmath} is a
PPO-style~\citep{schulman2017ppo} policy-gradient algorithm that replaces
the value network with a within-prompt baseline. For each prompt $x_i$,
it samples $n$ rollouts
$\{y_{i,k}\}_{k=1}^{n} \sim \pi_{\theta_{\mathrm{old}}}(\cdot \mid x_i)$,
receives verifier rewards $\{r_{i,k}\}_{k=1}^{n}$, and forms the
group-relative advantage
\begin{equation}
A^{\mathrm{GRPO}}_{i,k} \;=\;
\frac{r_{i,k} - \mu_i}{\sigma_i + \epsilon},
\qquad
\mu_i = \tfrac{1}{n}\!\sum_{k'} r_{i,k'},\;\;
\sigma_i = \mathrm{std}(\{r_{i,k'}\}).
\label{eq:grpo-adv}
\end{equation}
The policy is then updated with a clipped PPO surrogate and a KL penalty
to a reference policy,
\begin{equation}
\mathcal{L}_{\mathrm{GRPO}}(\theta) \;=\;
-\,\mathbb{E}_{i,k,t}\!\left[
\min\!\big(\rho_{i,k,t}(\theta)\, A^{\mathrm{GRPO}}_{i,k},\;
\mathrm{clip}(\rho_{i,k,t}(\theta), 1\!-\!\varepsilon, 1\!+\!\varepsilon)\,
A^{\mathrm{GRPO}}_{i,k}\big)
\right]
+ \beta\, D_{\mathrm{KL}}\!\left(\pi_\theta \Vert \pi_{\mathrm{ref}}\right),
\label{eq:grpo-loss}
\end{equation}
where
$\rho_{i,k,t}(\theta) = \pi_\theta(y_{i,k,t}\mid x_i, y_{i,k,<t}) /
\pi_{\theta_{\mathrm{old}}}(y_{i,k,t}\mid x_i, y_{i,k,<t})$ and
$\varepsilon$ is the clip range. A known weakness of
Equation~\eqref{eq:grpo-adv} is the zero-variance case. When every
rollout for a prompt receives the same reward, most commonly when all rollouts are wrong, and then the prompt contributes no gradient.

\vspace{-3mm}

\begin{figure}
    \centering
    \includegraphics[width=1.0\linewidth]{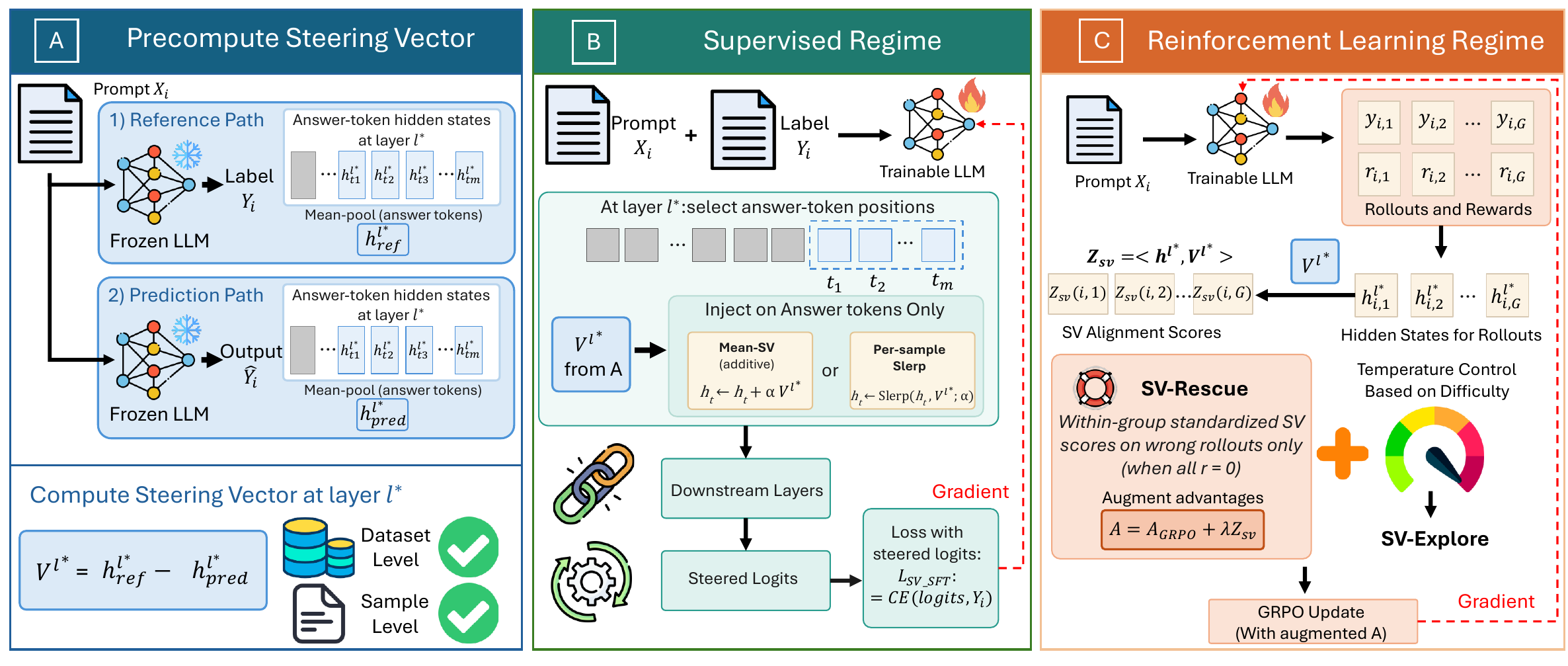}
    \caption{Overall framework. Panel (A) precomputes the off-policy displacement vector, which serves as the injection signal for the Supervised Regime in (B). In parallel, the Reinforcement Learning Regime in (C) applies the steering mechanism independently to rescue zero-variance rollout groups.}
    \vspace{-5mm}
    \label{fig:framework}
\end{figure}

\section{Steering Vector Post-training: Supervised Regime}
\label{sec:method-sft}

\subsection{Method}
\label{subsec:method-sft}

Since the source of the problem above is that teacher-forced training is off-policy, our goal is to construct a steering direction whose injection during teacher-forced training makes the labels appear less off-policy to the policy under update. Following the framework introduced in Section~\ref{sec:intro}, we describe the hidden-state-hook regime along three axes, namely the \emph{direction} we use, the \emph{placement} where we inject it, and the \emph{operation} that combines it with the existing hidden state.

\noindent \textbf{Direction.} To close the off-policy gap during teacher forcing, we need a direction that points from where the model's own activations naturally sit to where the reference activations sit. Reading this gap directly off the residual stream gives us such a direction by construction, and avoids any extra learning. Concretely, before training we sample one greedy completion $y_i^{\mathrm{Pred}}$ from the initial model on each training prompt $x_i$, while the reference completion $y_i^{\mathrm{GT}}$ is read from the dataset (in SFT) or produced by an external teacher (in distillation). Fixing a layer $\ell^\star$, we then run teacher forcing on both completions to read out the answer-token-pooled hidden states
\begin{equation}
h^{\ell^\star}_{\mathrm{ref}}(x_i) \;=\; \bar h^{\ell^\star}\!\left(x_i,\, y_i^{\mathrm{GT}}\right),
\qquad
h^{\ell^\star}_{\mathrm{pred}}(x_i) \;=\; \bar h^{\ell^\star}\!\left(x_i,\, y_i^{\mathrm{Pred}}\right).
\label{eq:method-href-hpred}
\end{equation}
The prompt-wise direction $\delta_i = h^{\ell^\star}_{\mathrm{ref}}(x_i) - h^{\ell^\star}_{\mathrm{pred}}(x_i)$ is, by definition, the off-policy gap measured at layer $\ell^\star$ on prompt $x_i$. We use either $\delta_i$ directly to retain heterogeneity across prompts, or its dataset-level mean:
\begin{equation}
v^{\ell^\star}
\;=\;
\mathbb{E}_i\!\left[
h^{\ell^\star}_{\mathrm{ref}}(x_i)
\;-\;
h^{\ell^\star}_{\mathrm{pred}}(x_i)
\right],
\label{eq:method-sv}
\end{equation}
which captures the direction that drives the off-policy gap on average. Both are computed once before training and fixed throughout, so the direction is not a moving target chasing the policy as it updates.

\noindent \textbf{Placement.} We inject at the middle layer $\ell^\star = L/2 + 1$, following prior steering-vector work that has consistently found mid-stack layers most effective for behavioral interventions~\citep{panickssery2024caa, turner2023activation}, and we use the same layer for both constructing and injecting the direction. We further restrict injection to answer-token positions because the loss is computed only on answer tokens, so the SV signal should be concentrated where it can flow into the gradient. Injecting on prompt tokens would corrupt the shared prompt encoding without contributing to the loss. During the forward pass of training, we therefore inject the residual stream at layer $\ell^\star$ and answer tokens only:

\begin{equation}
\tilde h^{\ell^\star}_t \;=\;
\begin{cases}
\mathrm{Inject}\!\left(h^{\ell^\star}_t,\, v^{\ell^\star}\right)
  & \text{if } m_{i,t}=1 \text{ (answer token)}, \\
h^{\ell^\star}_t
  & \text{otherwise},
\end{cases}
\label{eq:method-hook}
\end{equation}

\noindent \textbf{Operation.} The remaining choice is how to combine the direction with the existing hidden state. The most natural primitive is the additive operation of Equation~\eqref{eq:additive-inject}, $\tilde h = h + \alpha v$, which gives a clean linear correction along the direction at small $\alpha$. The additive primitive does inflate the norm of the residual stream as $\alpha$ grows, which can push the activation off the manifold the downstream layers were trained on, so we also test spherical linear interpolation (Slerp, Equation~\eqref{eq:slerp-def}) as a complementary primitive that rotates the hidden state toward the reference direction without changing its norm. We instantiate $\mathrm{Inject}(\cdot, \cdot)$ with both. \emph{Mean-SV} uses the additive primitive with the global $v^{\ell^\star}$ from Equation~\eqref{eq:method-sv}, and \emph{Per-sample Slerp} uses the spherical primitive and substitutes the per-prompt $\delta_i$ for $v^{\ell^\star}$. The steered hidden state propagates through the downstream layers to the logits, and the SFT objective becomes:
\begin{equation}
\mathcal{L}_{\mathrm{SV\text{-}SFT}}(\theta)
\;=\;
-\sum_{i}\sum_{t}
m_{i,t}\,
\log \pi_\theta^{\mathrm{SV}}\!\left(
y_{i,t}\,\big|\, x_i,\, y_{i,<t}
\right),
\label{eq:method-loss}
\end{equation}
where $\pi_\theta^{\mathrm{SV}}$ denotes the model whose residual stream at block $\ell^\star$ has been steered by Equation~\eqref{eq:method-hook}, and the gradient flows through $\tilde h^{\ell^\star}$ back to all trainable parameters. $\mathcal{L}_{\mathrm{SV\text{-}SFT}}$ collapses to $\mathcal{L}_{\mathrm{SFT}}$ when the injected direction is zero. The same procedure transfers verbatim to distillation, with the reference completion now produced by an external teacher. The full framework is shown in \cref{fig:framework} (A) and (B).

\noindent \textbf{Does the SV leak target information?} $\delta_i$ summarizes the full reference and is injected at every answer position, so it gives each position information that plain teacher forcing does not, which is the intended training signal. It is not evaluation leakage, because the SV is built only from training-set references, and \emph{no vector is injected at test time}, so any gain must be stored in the parameters.

% =====================================================================
\subsection{Supervised Fine-Tuning}
\label{subsec:sft}

\subsubsection{Experimental Setting}
\label{subsubsec:sft-setting}

\noindent \textbf{Models and datasets.} We train four models that span family and scale, namely Qwen2.5-1.5B-Instruct and Qwen2.5-7B-Instruct~\citep{qwen2025technical}, Qwen3-4B (no-thinking mode)~\citep{yang2025qwen3}, and llama3.2-3B~\cite{grattafiori2024llama}, on four datasets that span grade-school math (GSM8K)~\citep{cobbe2021gsm8k}, competition-level math (MATH)~\citep{hendrycks2021math}, code synthesis (MBPP)~\citep{austin2021mbpp}, and discrete reasoning (StrategyQA)~\citep{geva2021strategyqa}.

\noindent \textbf{Training and evaluation protocol.} Following common parameter-efficient fine-tuning practice for LLMs~\cite{xu2025parenting} ,all runs use LoRA~\citep{hu2022lora} with rank $r=8$, learning rate $5\times 10^{-5}$, batch size $8$, and one epoch. Steering injection uses Per-sample Slerp at $\alpha=0.1$, Mean-SV at $\alpha=0.5$ and layer $\ell^\star = L/2 +1$ over the full answer span. All evaluations (including results in other parts as well) use vLLM with greedy decoding unless otherwise specified. All numbers reported below are accuracies on the official test split. Experiments are done using 2 Nvidia L40-S.

\noindent \textbf{Hyperparameter selection.} The layer $\ell^\star=L/2+1$ is fixed in advance by the mid-stack rule of inference-time steering~\citep{panickssery2024caa}. The operation and $\alpha$ are selected once on Qwen2.5-1.5B with GSM8K, training on $6{,}000$ training examples and validating on the remaining ones, and this single configuration is reused unchanged for every model and dataset.

\subsubsection{Main Results}
\label{subsubsec:sft-main}

To comprehensively compare normal SFT and SFT with steering vector, we show the results for four different models and four different datasets in \cref{tab:sft-main}. The results show that using SV during SFT improves the performance after SFT in general, and we make the following observations:

\begin{wraptable}[15]{r}{0.5\textwidth}
\centering
\vspace{-4mm}
\caption{\small Comparison of test accuracy (\%) after SFT across four models and datasets. The best result per row is shown in \textbf{bold}. SV-augmented SFT improves over plain SFT in general.}
\label{tab:sft-main}
\small
\resizebox{0.5\textwidth}{!}{
\begin{tabular}{llccc}
\toprule
Model & Dataset & Plain & Mean-SV & Per-sample Slerp \\
\midrule
\multirow{4}{*}{Qwen2.5-1.5B}
 & GSM8K       & 59.14          & 62.85          & \textbf{64.14} \\
 & MATH        & 24.60          & 27.00          & \textbf{27.20} \\
 & MBPP        & \textbf{47.00} & 45.80          & 45.60 \\
 & StrategyQA  & 62.74          & \textbf{64.48} & 63.61 \\
\midrule
\multirow{4}{*}{Qwen2.5-7B}
 & GSM8K       & 76.88          & 80.14          & \textbf{80.21} \\
 & MATH        & 43.20          & 44.00          & \textbf{47.40} \\
 & MBPP        & 64.80          & 65.20          & \textbf{65.20} \\
 & StrategyQA  & 78.17          & 79.62          & \textbf{79.91} \\
\midrule
\multirow{4}{*}{Qwen3-4B}
 & GSM8K       & 80.06          & 81.73          & \textbf{84.61} \\
 & MATH        & 44.80          & \textbf{47.80} & 47.60 \\
 & MBPP        & 57.40          & \textbf{59.60} & 59.40 \\
 & StrategyQA  & 71.62          & 71.18          & \textbf{72.93} \\
\midrule
\multirow{4}{*}{Llama-3.2-3B}
 & GSM8K       & 62.09          & \textbf{63.08} & 61.79 \\
 & MATH        & 25.00          & 25.00          & \textbf{27.40} \\
 & MBPP        & 48.20          & \textbf{48.80} & 48.60 \\
 & StrategyQA  & \textbf{66.23} & 65.07          & 64.63 \\
\bottomrule
\end{tabular}
}
\vspace{-5mm}
\end{wraptable}

\noindent $\bullet$ Adding a steering vector during SFT is positive over Plain on the three Qwen models in $11$ of $12$ cells. The only negative result among Qwen models is Qwen2.5-1.5B on MBPP, and we analyze the cause in Appendix \cref{sec:sft_analysis_appendix}.

\noindent $\bullet$ Per-sample Slerp generally beats Mean-SV. Per-sample Slerp wins in $8$ of $12$ Qwen cells over Mean-SV despite using a $5\times$ smaller injection coefficient.

\noindent $\bullet$ The default configuration transfers poorly to Llama-3.2-3B. Per-sample Slerp gives a mean gain of only $+0.23$pp across the four Llama datasets. We analyze this in detail in the following analysis section, and the results indicate the failure is due to layer choice rather than the SV mechanism itself.

\subsubsection{Analysis}
\label{subsubsec:sft-analysis}

We run a series of ablations on Qwen2.5-1.5B with GSM8K to understand what drives the gain.

% \noindent \textbf{Influence of $\alpha$ and the injection layer.} To understand the dynamics of SFT with SV, we sweep the injection coefficient $\alpha$ for both Per-sample Slerp and Mean-SV additive at fixed $\ell^\star=15$ and the layer for Per-sample Slerp at fixed $\alpha=0.1$. As is shown in \cref{fig:sft-sensitivity} (a), on Qwen2.5-1.5B, Per-sample Slerp collapses to $0\%$ at $\alpha=1.0$ because the hidden state is fully rotated away from the prediction direction, while Mean-SV additive has a broad plateau. Per-sample Slerp therefore achieves a higher peak but is more sensitive to the choice of $\alpha$.  \hua{what is the relationship between (a) and (b)? why putting them together in one paragraph? }As is shown in \cref{fig:sft-sensitivity} (b), Llama-3.2-3B has a strikingly different layer profile compared to Qwen: $\ell^\star \in \{10, 13, 19\}$ all surpass Plain SFT and $\ell^\star=10$ and $\ell^\star=19$ are tied at the peak. The negative Llama row at the default $\ell^\star=15$ in \cref{tab:sft-main} is therefore not a failure of SV on Llama but a failure of layer transfer across families, and the optimal injection layer is model-specific.

\begin{figure}[h!]
\centering
\vspace{-3mm}
\begin{subfigure}[t]{0.32\linewidth}
  \centering
  \includegraphics[width=\linewidth]{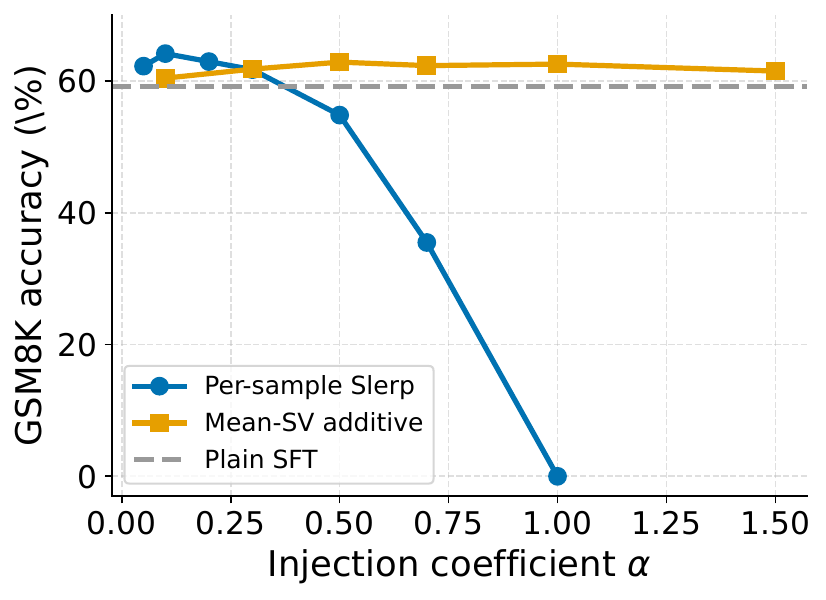}
  \vspace{-6mm}
  \caption{\small $\alpha$-sensitivity curves of Per-sample Slerp and Mean-SV at fixed $\ell^\star=15$. Slerp peaks sharply while Mean-SV has a broad plateau.}
  \label{fig:sft-sensitivity:a}
\end{subfigure}\hfill
\begin{subfigure}[t]{0.32\linewidth}
  \centering
  \includegraphics[width=\linewidth]{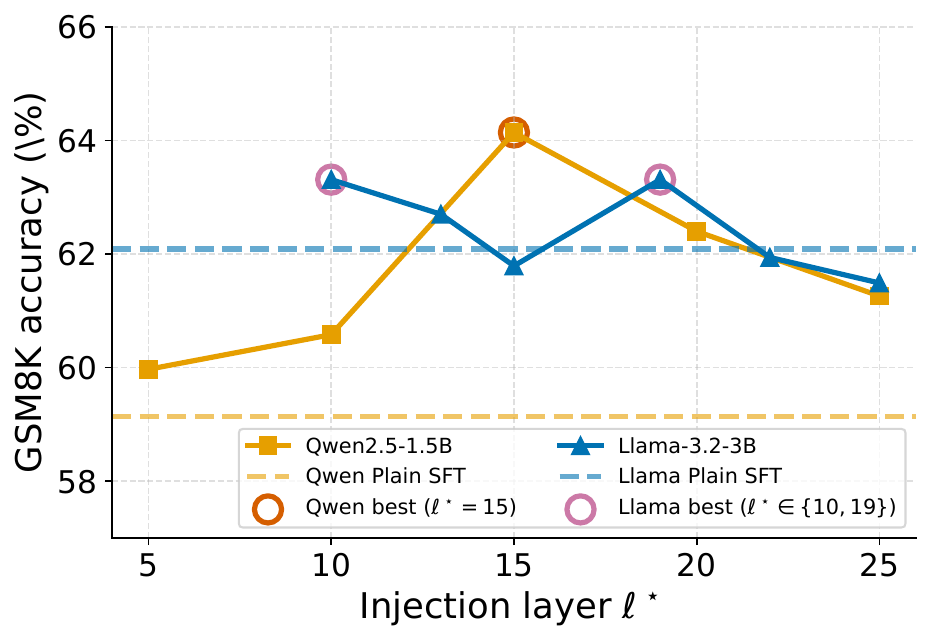}
  \vspace{-6mm}
  \caption{\small Accuracy of Per-sample Slerp as the injection layer $\ell^\star$ is varied with $\alpha=0.1$. The best layer for Qwen and Llama is quite different.}
  \label{fig:sft-sensitivity:b}
\end{subfigure}\hfill
\begin{subfigure}[t]{0.32\linewidth}
  \centering
  \includegraphics[width=\linewidth]{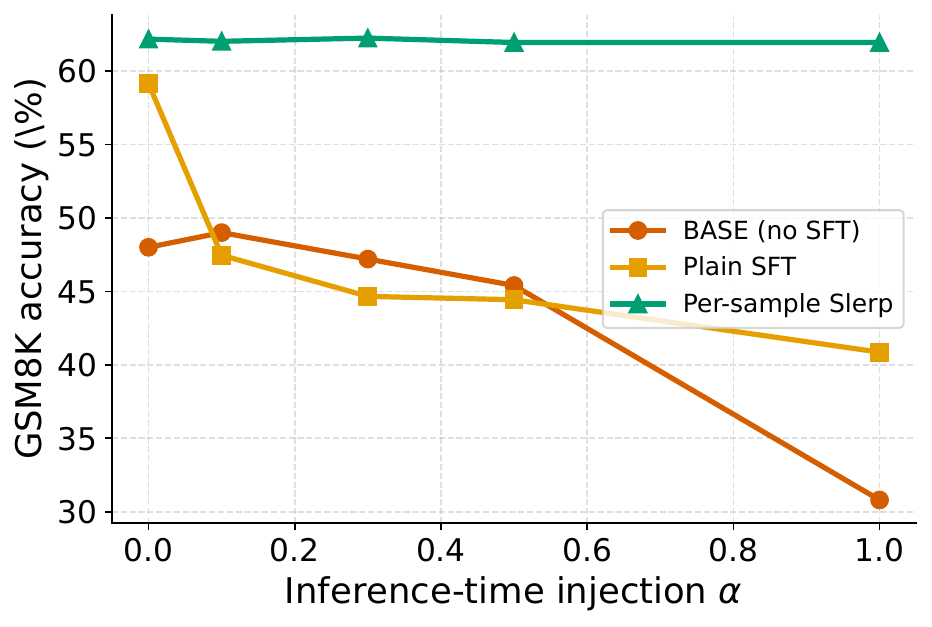}
  \vspace{-6mm}
  \caption{\small Accuracy under inference-time SV injection on three checkpoints, showing that adding SV at inference time does not help.}
  \label{fig:sft-sensitivity:c}
\end{subfigure}
\caption{Analysis of our SV-augmented SFT on Qwen2.5-1.5B with GSM8K.}
\vspace{-2mm}
\label{fig:sft-sensitivity}
\end{figure}

\noindent \textbf{Sensitivity to the injection coefficient $\alpha$.} To understand the dynamics of SFT with SV, we first sweep the injection coefficient $\alpha$ for both Per-sample Slerp and Mean-SV additive at a fixed layer $\ell^\star=15$. As is shown in \cref{fig:sft-sensitivity} (a), on Qwen2.5-1.5B, Per-sample Slerp collapses to $0\%$ at $\alpha=1.0$ because the hidden state is fully rotated away from the prediction direction, while Mean-SV additive has a broad plateau. Per-sample Slerp therefore achieves a higher peak but is more sensitive to the choice of $\alpha$. This suggests that Slerp unlocks superior performance provided that $\alpha$ is carefully tuned to avoid representation collapse, whereas Mean-SV serves as a safer, more forgiving alternative when extensive hyperparameter search is prohibitive.

\noindent \textbf{Influence of injection layer $\ell^\star$.} Next, we investigate the influence of the injection layer by fixing $\alpha=0.1$ for Per-sample Slerp. As is shown in \cref{fig:sft-sensitivity} (b), Llama-3.2-3B has a strikingly different layer profile compared to Qwen: $\ell^\star \in \{10, 13, 19\}$ all surpass Plain SFT, and $\ell^\star=10$ and $\ell^\star=19$ are tied at the peak. The negative Llama row at the default $\ell^\star=15$ in \cref{tab:sft-main} is therefore not a failure of SV on Llama but a failure of layer transfer across families, indicating that the optimal injection layer or placement choice is highly model-specific. What transfers from inference-time practice is the mid-stack region, since on both families neither shallow nor deep layers are ever optimal. For a new backbone, a $0.25$-epoch layer sweep validated on the last $500$ training examples recovers the best layer on both Qwen2.5-1.5B and Llama-3.2-3B without test data, as detailed in \cref{app:hparam}.

\begin{wraptable}[13]{r}{0.5\textwidth}
\centering
\vspace{-4mm}
\caption{\small Four representative SV injection variants for SV SFT on Qwen2.5-1.5B with GSM8K. The full set of variants we tested is reported in Appendix~\cref{sec:sft_analysis_appendix}. The best result is shown in \textbf{bold}. Per-sample Slerp outperforms every alternative SV construction we tested.}
\label{tab:sft-variants}
\small
\resizebox{0.5\textwidth}{!}{
\begin{tabular}{p{4cm}ccc}
\toprule
Method & $\alpha$ & Accuracy & $\Delta$ \\
\midrule
Plain LoRA (no SV)                                & --             & 59.14 & 0 \\
Random per-sample Gaussian (control)              & 0.50           & 54.82 & $-4.32$ \\
PCA SV                                        & 0.50           & 59.29 & $+0.15$ \\
Multi-layer (L=10/15/20)                 & 0.50/0.50/0.50 & 60.85 & $+1.71$ \\
Multi-layer (L=10/15/20)                   & 0.10/0.50/0.10 & 62.62 & $+3.48$ \\
Per-sample additive                               & 0.50           & 62.93 & $+3.79$ \\
\textbf{Per-sample Slerp (full answer)}           & \textbf{0.10}  & \textbf{64.14} & $\mathbf{+5.00}$ \\
\bottomrule
\end{tabular}
}
\vspace{-5mm}
\end{wraptable}

\noindent \textbf{Direction and injection variants.} To check that our findings are generalizable, we compare our method against four representative alternatives in~\cref{tab:sft-variants} and defer the full set of variants we tested to Appendix~\cref{sec:appendix-variants}. \emph{Random per-sample Gaussian} replaces $\delta_i$ with a Gaussian noise and serves as a control for whether SV causes the improvement. \emph{PCA SV}~\cite{zou2023representation} replaces the dataset-level mean with the principal component (first 5) of the per-sample $\delta_i$ matrix to reduce the influence of confounding. \emph{Multi-layer} inject Mean-SV at three transformer blocks $L \in \{10, 15, 20\}$ simultaneously. \emph{Per-sample additive} keeps the per-sample $\delta_i$ but injects it through the additive primitive of Equation~\eqref{eq:additive-inject} rather than the spherical primitive of Equation~\eqref{eq:slerp-def}, isolating the contribution of the Slerp injection geometry from the per-sample direction. Overall, we have the following conclusions:

\noindent $\bullet$ The random Gaussian control hurts the model, which shows that what our method exploits is the specific GT-Pred direction in residual space and not the existence of an additive perturbation.

\noindent $\bullet$ PCA gains little over Plain LoRA, indicating that confounding is not a major problem here.

\noindent $\bullet$ Multi-layer Mean-SV is sensitive to the $\alpha$ schedule. The uniform $0.50/0.50/0.50$ schedule underperforms the single-layer Mean-SV, while the tuned schedule that concentrates the weight on L15 beats it. Multi-layer injection can therefore help, but only with careful per-layer tuning.

\noindent $\bullet$ Per-sample additive clearly improves but still trails Per-sample Slerp, which decomposes the headline gain into per-sample heterogeneity and the spherical injection geometry to show that both ingredients are needed.

\noindent \textbf{Inference-time SV is qualitatively different from training-time SV.} The original use of SVs is at inference time, and we test the same direction at inference, with results in (c) of \cref{fig:sft-sensitivity}.  This experiment serves two purposes. First, the base and Plain-SFT curves form a control test. If our direction were simply a good inference-time steer, injecting it into a model that never saw it during training should already help. It does not, so the training-time gain is not a post-hoc steering effect. Second, the SV-trained curve forms a distribution-match test. Since this checkpoint is trained with the SV on every answer token, keeping the SV at inference matches its training distribution, yet it performs worse than dropping the SV. We attribute this to the SV offloading the style and format component of the reference shift during training. In support of this account, the change that training makes to the layer-$\ell^\star$ hidden state is nearly orthogonal to the SV (cosine $\approx -0.05$), whereas SV-free training aligns positively with it, so the gradient concentrates on the correctness-relevant remainder and re-injecting the SV at inference only perturbs the hidden state. Results on more models and datasets are in \cref{app:train-vs-inference}.

% =====================================================================
\subsection{Distillation}
\label{subsec:distill}

\subsubsection{Experimental Setting}
\label{subsubsec:distill-setting}

\noindent \textbf{Teachers, students, and datasets.} We use three students and four datasets from \cref{subsubsec:sft-setting} and pair them with four teachers chosen to vary both mode and family, namely Qwen3-4B-Instruct-2507 in non-thinking mode and Qwen3-8B in thinking mode~\citep{qwen2025technical}, and Gemma-4-26B-A4B~\cite{farabet2026gemma4} in non-thinking and thinking mode, a different-family model. Each teacher is asked to produce one completion per training prompt, and we keep only the completions that pass the dataset's verifier.

\noindent \textbf{Distillation protocol.} The student is trained with the same LoRA configuration as in Section~\ref{subsubsec:sft-setting}. The steering vector is now constructed from the contrast between the student's hidden state on the teacher's completion, and the student's hidden state on its own completion of the same prompt.

\vspace{-2mm}

\subsubsection{Main Results}
\label{subsubsec:distill-main}

To comprehensively compare distillation and distillation with SV, we show the results for three different models in \cref{tab:distill-main}. We make the following observations:

\begin{table}[t]
\centering
\caption{Comparison of test accuracy (\%) across three students, four teachers, and four datasets. The best result is shown in \textbf{bold}. SV-augmented distillation improves over plain distillation on math, with the largest gains on thinking-mode teachers.}
\label{tab:distill-main}
\scriptsize
\setlength{\tabcolsep}{4pt}
\resizebox{0.95\textwidth}{!}{%
\begin{tabular}{llcccccccccccc}
\toprule
\multirow{2}{*}{Student} & \multirow{2}{*}{Dataset}
& \multicolumn{3}{c}{Qwen3-4B-Instruct}
& \multicolumn{3}{c}{Qwen3-8B}
& \multicolumn{3}{c}{Gemma-4-26B (no-thinking)}
& \multicolumn{3}{c}{Gemma-4-26B (thinking)} \\
\cmidrule(lr){3-5}\cmidrule(lr){6-8}\cmidrule(lr){9-11}\cmidrule(lr){12-14}
& & Plain & Mean-SV & Slerp
  & Plain & Mean-SV & Slerp
  & Plain & Mean-SV & Slerp
  & Plain & Mean-SV & Slerp \\
\midrule
\multirow{4}{*}{Qwen2.5-1.5B}
 & GSM8K       & \textbf{74.83} & \textbf{74.83}          & 74.68          & 65.81          & \textbf{67.55} & 67.40 & 70.58          & \textbf{71.11} & 70.43          & 49.58          & 60.42          & \textbf{60.50} \\
 & MATH        & 41.60          & 41.40 & \textbf{42.20} & 21.20          & 31.20          & \textbf{33.60} & 36.60          & \textbf{37.60}          & 37.40 & 11.80          & \textbf{15.00} & 14.00          \\
 & MBPP        & 49.00          & 48.80          & \textbf{50.00} & \textbf{46.00} & 45.60 & 45.40          & 48.80          & 48.00          & \textbf{48.80}          & 27.00          & 31.80          & \textbf{37.40} \\
 & StrategyQA  & 56.91 & \textbf{58.08}          & 55.31          & 55.31          & 59.83          & \textbf{60.84} & \textbf{61.86} & 60.99          & 58.52          & 20.82          & 19.07          & \textbf{21.11} \\
\midrule
\multirow{4}{*}{Qwen3-4B}
 & GSM8K       & 87.87          & \textbf{88.02}          & 87.95 & 76.42 & 75.51          & \textbf{78.62}          & 86.88          & \textbf{87.34} & 86.73          & 62.70          & 52.77          & \textbf{67.63} \\
 & MATH        & \textbf{67.60} & 67.20 & 67.00          & 66.00          & 65.80 & \textbf{66.20} & 66.00          & 64.80          & \textbf{67.80} & 19.40 & \textbf{27.20}          & 26.60          \\
 & MBPP        & 65.00          & \textbf{66.40} & \textbf{66.40} & \textbf{65.40} & 65.20 & \textbf{65.40}          & \textbf{65.20}          & 65.00          & 65.00          & 64.00          & \textbf{64.40} & \textbf{64.40} \\
 & StrategyQA  & \textbf{65.21} & 64.77          & 64.48          & 62.01          & 62.88          & \textbf{65.50} & 71.18          & \textbf{72.20} & 69.14          & 45.41          & \textbf{50.51} & 39.88          \\
\midrule
\multirow{4}{*}{Llama-3.2-3B}
 & GSM8K       & \textbf{77.48} & 76.72          & 76.65          & 67.17          & \textbf{68.23} & 67.85          & 71.42          & 72.63          & \textbf{72.93} & \textbf{47.61} & 45.79          & 46.63          \\
 & MATH        & 38.00          & 38.00          & 38.00          & 9.80           & \textbf{21.00} & 19.60          & 34.20          & 34.40          & \textbf{36.00} & 12.20          & 15.40          & \textbf{17.00} \\
 & MBPP        & \textbf{49.40}          & \textbf{49.40}          & 49.20          & 44.60          & \textbf{45.60} & 45.40          & 50.40          & 50.20          & \textbf{50.60} & 47.20          & \textbf{47.60} & 46.80          \\
 & StrategyQA  & 59.68          & 60.55          & \textbf{62.45} & \textbf{64.19} & 60.41          & 62.59          & 65.79          & 65.79          & \textbf{67.98} & \textbf{34.50} & 33.77          & 33.77          \\
\bottomrule
\end{tabular}%
}
\vspace{-6mm}
\end{table}

\noindent $\bullet$ Teacher mode dominates teacher family in determining the SV gain. Both thinking teachers (Qwen3-8B and Gemma-thinking) yield the biggest gains over Plain, up to $12\%$, while both no-thinking teachers are essentially flat. A plausible reason is that no-thinking teachers, even from a different family, produce outputs whose style is close to students, so the contrast between the teacher's completion and the student's own completion is small, and the SV has little to correct. Thinking teachers, by contrast, produce long reasoning traces that the student would hardly emit, which widens the contrast between the two populations and gives the SV a much richer correction signal.

\noindent $\bullet$ Llama-3.2-3B under-uses Per-sample Slerp under all four teachers, with mean SV gains well below the corresponding Qwen-student gains on every teacher. We attribute this to the same layer-choice mismatch identified in the SFT main table, not a failure of the SV mechanism itself.

\vspace{-2mm}

\subsubsection{Analysis}
\label{subsubsec:distill-analysis}
This subsection presents the two analyses that directly support our observation about the distillation results. Further analyses, including normalized SV vs.\ unnormalized SV, are deferred to \cref{sec:distill_analysis_appendix}.

\begin{figure}[h]
\centering
\vspace{-5mm}
\begin{subfigure}[t]{0.32\linewidth}
  \centering
  \includegraphics[width=\linewidth]{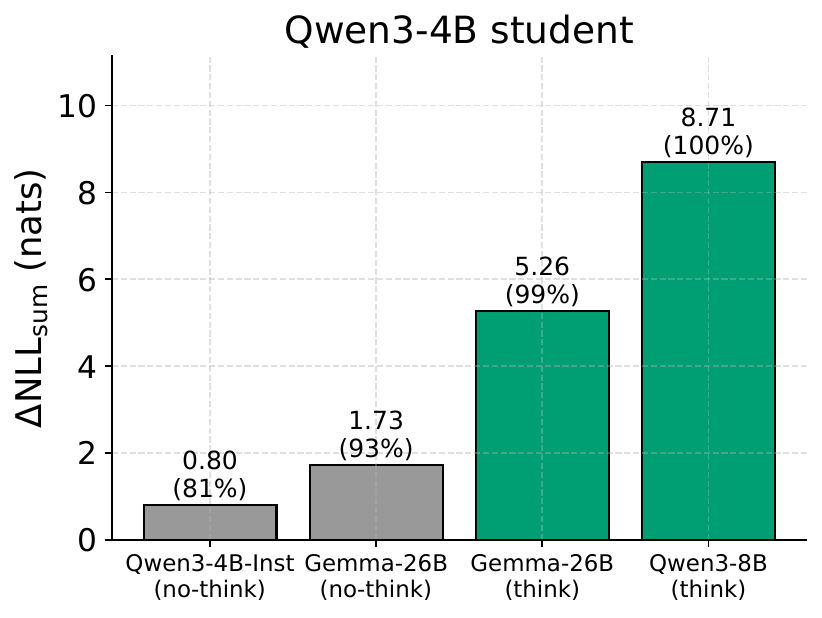}
  \vspace{-5mm}
  \caption{\small $\Delta\mathrm{NLL}_{\mathrm{sum}}$ on Qwen3-4B across four distillation teachers. Numbers in parentheses are $\mathrm{frac}(\Delta\mathrm{NLL}_{\mathrm{sum}} > 0)$ for samples.}
  \label{fig:distill-analysis:a}
\end{subfigure}\hfill
\begin{subfigure}[t]{0.32\linewidth}
  \centering
  \includegraphics[width=\linewidth]{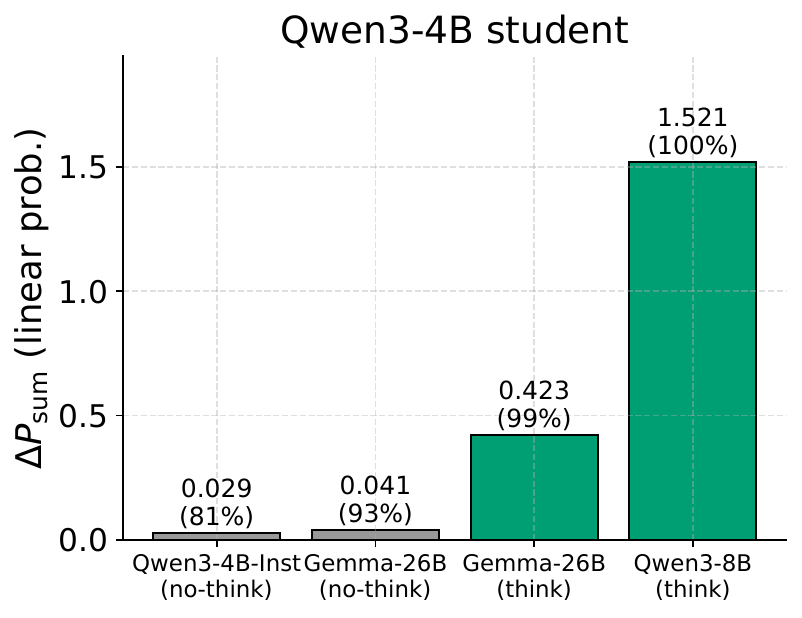}
  \vspace{-5mm}
  \caption{\small $\Delta P_{\mathrm{sum}}$ on the same four cells, in linear probability units. The ordering matches figure (a).}
  \label{fig:distill-analysis:b}
\end{subfigure}\hfill
\begin{subfigure}[t]{0.32\linewidth}
  \centering
  \includegraphics[width=\linewidth]{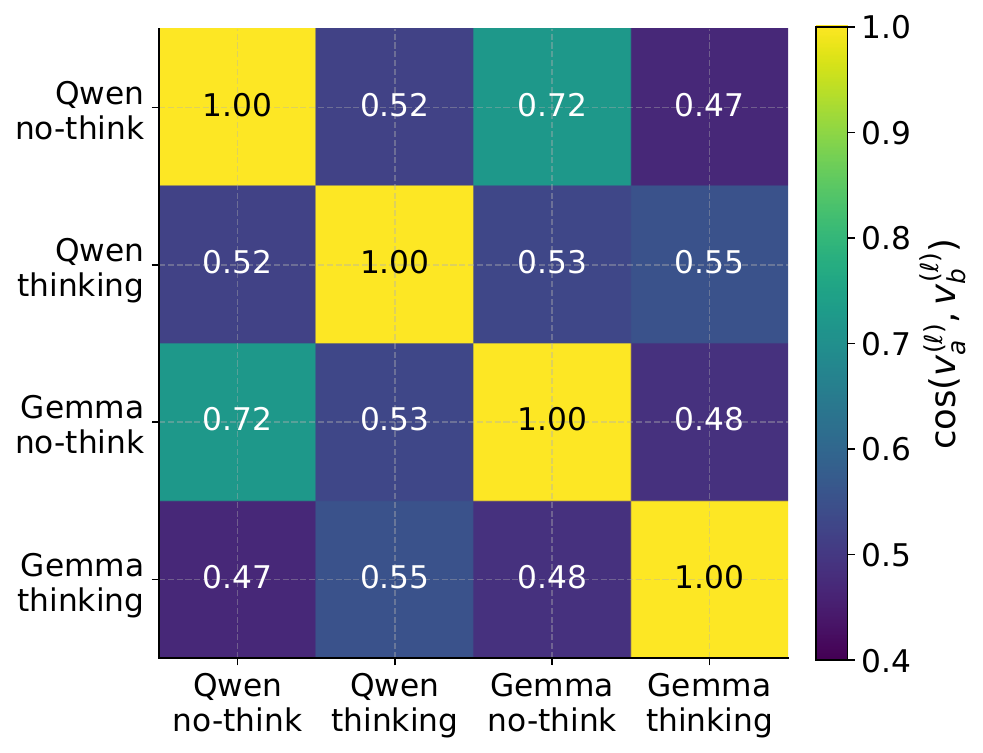}
  \vspace{-5mm}
  \caption{\small Cosine similarity of the SV direction $v^{(\ell)}$ between the four teachers, averaged over the 12 (student, dataset) pairs of each teacher.}
  \label{fig:distill-teacher}
\end{subfigure}
  \vspace{-1mm}
\caption{Analysis on distillation settings. SV-augmented distillation closes the off-policy gap, and the SV direction is reshaped more by switching teacher mode than by switching teacher family.}
\label{fig:distill-analysis}
\vspace{-7mm}
\end{figure}

\noindent \textbf{Mitigating the off-policy gap.} If the SV truly closes the off-policy gap, then injecting it into the base model under teacher forcing should raise the probability of the reference completion without any parameter update. We apply the training-time SV hook on $500$ training samples and report two statistics on all answer tokens by teacher forcing. The first, $\Delta\mathrm{NLL}_{\mathrm{sum}}$, measures the log-probability gain on the full label sequence. The second, $\Delta P_{\mathrm{sum}}$, measures the linear probability mass moved onto label tokens. Both metrics tell the same story in \cref{fig:distill-analysis:a,fig:distill-analysis:b}:  the evidence is consistent with gap mitigation, with large and consistent effects on thinking-teacher cells and much weaker effects on non-thinking-teacher cells. The full ten-cell sweep in \cref{sec:distill_analysis_appendix} shows that the cells without activation-level mitigation are exactly the cells whose SV-distill accuracy gain is small.

\noindent \textbf{Influence of teacher mode and family.} The cross-teacher cosine similarity shown in~\cref{fig:distill-teacher} provides the geometric explanation of the main-text observation that mode dominates family. Same-mode teachers from different families remain well-aligned in the SV direction, while different-mode pairs align poorly even within the same family. Switching the teacher's mode, therefore, reshapes the SV direction far more than switching the teacher's family does, which is consistent with the results pattern that teacher mode dominates teacher family in determining the SV gain.

\subsection{Multi-Seed Reliability}
\label{subsec:seeds}

Since the main tables use a single seed, we rerun representative SFT and distillation cells with seeds $\{42, 1, 2\}$, using the same pipeline and the fixed configuration. As shown in \cref{tab:seeds}, every reasoning and distillation cell improves over Plain on all three seeds. The Qwen2.5-1.5B GSM8K gain is about eight times the larger seed standard deviation, and distillation from the thinking teacher nearly doubles plain distillation, while MBPP stays neutral within noise, consistent with the reference direction being the wrong target for code. 

\begin{table}[t]
\centering
\caption{Accuracy (\%, mean $\pm$ std over three seeds) of Plain and Per-sample Slerp. ``Distill'' uses the Qwen3-8B thinking teacher.}
\label{tab:seeds}
\resizebox{\textwidth}{!}{%
\begin{tabular}{lccccccc}
\toprule
 & 1.5B GSM8K & 7B GSM8K & 1.5B MATH & Qwen3-4B MATH & 7B MATH & 1.5B MATH (Distill) & 1.5B MBPP \\
\midrule
Plain            & $58.81 \pm 0.37$ & $77.66 \pm 1.73$ & $25.80 \pm 0.53$ & $42.47 \pm 2.20$ & $42.47 \pm 1.22$ & $18.13 \pm 2.30$ & $46.13 \pm 0.31$ \\
Per-sample Slerp & $62.75 \pm 0.50$ & $80.74 \pm 0.49$ & $27.00 \pm 0.20$ & $47.07 \pm 1.70$ & $46.20 \pm 1.40$ & $33.47 \pm 1.14$ & $46.47 \pm 1.01$ \\
$\Delta$         & $+3.94$ & $+3.08$ & $+1.20$ & $+4.60$ & $+3.73$ & $+15.34$ & $+0.34$ \\
\bottomrule
\end{tabular}}
\vspace{-6mm}
\end{table}

\section{Steering Vector Post-training: Reinforcement Learning Regime}
\label{sec:method-rl}

\subsection{Method}
\label{subsec:method-rl}

As reviewed in Section~\ref{subsec:grpo}, GRPO collapses on zero-variance groups. One case in practice is the all-wrong group. These groups concentrate on the hardest prompts, exactly where learning is most needed, and the training budget is therefore wasted on prompts that produce zero gradient. In this section, we use advantage-repair to fix this problem. Following the framework introduced in Section~\ref{sec:intro}, we describe the advantage-repair regime along three axes as well.

\noindent \textbf{Direction.} To recover a learning signal on zero-variance groups, we need a continuous proxy for correctness that does not rely on the verifier. The natural choice is the residual-stream direction along which correct rollouts differ from wrong ones. Projecting a rollout onto this direction yields a continuous proxy score that is expected to correlate with the correct-vs-wrong contrast, even when the group-level GRPO advantage is uninformative. Concretely, before GRPO training, we run the reference model on the training prompts to collect a buffer of rollouts, label each as correct or wrong with the verifier, and define:
\begin{equation}
v^{\ell^\star}_{\mathrm{corr}}
\;=\;
\mathbb{E}_{(i,k) \in \mathrm{Correct}}\!\left[\bar h^{\ell^\star}_{i,k}\right]
\;-\;
\mathbb{E}_{(i,k) \in \mathrm{Wrong}}\!\left[\bar h^{\ell^\star}_{i,k}\right],
\label{eq:rl-sv-direction}
\end{equation}
the answer-token-pooled mean-activation difference at layer $\ell^\star$ between correct and wrong rollouts. Recomputing this direction online during RL would introduce a moving target as the policy shifts and would add a per-step encoding cost, so we precompute it once and freeze it.

\noindent \textbf{Placement.} Following the SFT setup, we use the middle layer $\ell^\star = L/2 +1$. The score for rollout $(i,k)$ is the average projection over the latter half $W_{i,k}$ of its answer tokens:
\begin{equation}
z^{\mathrm{sv}}_{i,k}
\;=\;
\frac{1}{|W_{i,k}|}
\sum_{t \in W_{i,k}}
\big\langle h^{\ell^\star}_{i,k,t},\, \hat v^{\ell^\star}_{\mathrm{corr}}\big\rangle,
\label{eq:rl-sv-score}
\end{equation}
Intuitively, rollouts whose hidden states align more closely with the correct-rollout pattern receive higher scores, so even within an all-wrong group, the score distinguishes.

\noindent \textbf{Operation.} The remaining choice is how to feed $z^{\mathrm{sv}}_{i,k}$ back into GRPO. The GRPO advantage separates correct from wrong rollouts but treats all wrong rollouts of a prompt identically, and on all-wrong groups it vanishes entirely. We therefore leave correct rollouts untouched and act on every rollout the verifier has scored as wrong. We standardize $z^{\mathrm{sv}}_{i,k}$ within each group to $\bar z^{\mathrm{sv}}_{i,k}$ to match the scale of the GRPO advantage, and shift the advantage on wrong rollouts only:
\begin{equation}
A_{i,k}
\;=\;
A^{\mathrm{GRPO}}_{i,k}
\;+\;
\lambda \cdot \bar z^{\mathrm{sv}}_{i,k} \cdot \mathbf{1}\!\left[r_{i,k}=0\right],
\label{eq:rl-sv-rescue}
\end{equation}
On all-wrong groups, $A^{\mathrm{GRPO}}_{i,k}=0$ and the advantage reduces to the SV ranking, turning an otherwise silent prompt into a learning signal; in mixed-reward groups, the term additionally ranks the wrong rollouts by their alignment with the correctness direction. We call this advantage modification \emph{SV-Rescue}. Because it touches only the advantage estimator, it composes cleanly with orthogonal rollout-side mechanisms. For example, the sampling temperature can be raised after steps that produced many all-wrong groups, encouraging more exploration on hard prompts. We call SV-Rescue applied on top of such a temperature schedule \emph{SV-Explore}. The full framework is shown in \cref{fig:framework} (C).

\begin{table}[t]
\centering
\begin{minipage}[t]{0.48\linewidth}
\centering
\caption{\small Comparison of final-step validation accuracy (\%) for GRPO post-training. The best result per row is shown in \textbf{bold}. SV-augmented GRPO consistently beats Plain GRPO on the math datasets.}
\label{tab:rl-main}
\small
\resizebox{\linewidth}{!}{
\begin{tabular}{llccc}
\toprule
Student & Training dataset & Plain & SV-Rescue & SV-Explore \\
\midrule
\multirow{3}{*}{Qwen2.5-1.5B}
 & GSM8K        & 77.72 & 78.35 & \textbf{79.18} \\
 & DeepScaleR   &   40.22  &  41.76 & \textbf{42.30}  \\
 & DeepCoder    & 13.98    & \textbf{16.13}    & 12.90    \\
\midrule
\multirow{3}{*}{Qwen3-4B}
 & GSM8K        & 90.76 & 91.04 & \textbf{91.99} \\
 & DeepScaleR   & 52.33    & 54.12    & \textbf{55.96}    \\
 & DeepCoder    &  \textbf{37.19}   &  35.18   &   36.18  \\
\midrule
\multirow{3}{*}{Llama-3.2-3B}
 & GSM8K        &  81.83    & 84.49    & \textbf{86.71}    \\
 & DeepScaleR   &  36.69   &  38.89   &   \textbf{38.92} \\
 & DeepCoder    & 21.05    & 21.05    & \textbf{24.21}    \\
\bottomrule
\end{tabular}
}
\end{minipage}\hfill
\begin{minipage}[t]{0.48\linewidth}
\centering
\caption{\small Out-of-distribution (OOD) evaluation accuracy (\%) of the GRPO checkpoints from \cref{tab:rl-main}. The best result is shown in \textbf{bold}. The in-domain SV gain transfers cleanly to OOD.}
\label{tab:rl-ood}
\small
\resizebox{\linewidth}{!}{
\begin{tabular}{llcccc}
\toprule
Student & Method & MATH-500 & AIME-24 & AIME-25 & LiveCodeBench \\
\midrule
\multirow{4}{*}{Qwen2.5-1.5B}
 & Plain      & 34.60 & 1.67 & 0.33 &  7.75 \\
 & SV-Rescue  & 35.80 & \textbf{3.33} & \textbf{1.00} & 7.75\\
 & SV-Explore & \textbf{37.00} & 2.33 & 0.67 & \textbf{8.25} \\
\midrule
\multirow{4}{*}{Qwen3-4B}
 & Plain      & 55.20 & 22.00 & 21.00 & 34.00 \\
 & SV-Rescue  & \textbf{56.80} & 23.33 & \textbf{22.66} & 33.00\\
 & SV-Explore & 56.40 & \textbf{24.66} & 22.00 & \textbf{36.00} \\
\midrule
\multirow{4}{*}{Llama-3.2-3B}
 & Plain      & 26.60 & 4.00 & 0.33 & 20.50 \\
 & SV-Rescue  & \textbf{32.00} & \textbf{4.33} & 0.67 & \textbf{22.50} \\
 & SV-Explore & 30.80 & \textbf{4.33} & \textbf{1.00} & 21.25 \\
\bottomrule
\end{tabular}
}
\end{minipage}
\vspace{-6mm}
\end{table}

\subsection{Experiments}
\label{subsec:rl}

\subsubsection{Experimental Setting}
\label{subsubsec:rl-setting}

\noindent \textbf{Models, datasets, and initial checkpoints.} We run RL on the same three models in \cref{subsec:distill}. Each student is paired with three training datasets that span easy math (GSM8K)~\citep{cobbe2021gsm8k}, hard math (DeepScaleR)~\citep{luo2025deepscaler}, and code (DeepCoder)~\citep{deepcoder2025}.

\noindent \textbf{GRPO configuration.} All runs use the verl~\citep{sheng2025hybridflow} GRPO implementation with $n=8$ rollouts per prompt, KL coefficient $\beta=5\times 10^{-3}$, base sampling temperature $T_{\mathrm{low}}=0.7$, learning rate $5\times 10^{-5}$. The actor is updated through LoRA with rank $r=8$ on the Qwen students, and through full-parameter fine-tuning on Llama-3.2-3B. The SV used by both SV-Rescue and SV-Explore is the same precomputed correctness direction using 500 samples in total. We use $\lambda = 0.1$ for our method.

\noindent \textbf{In-domain and out-of-distribution evaluation.} For each (student, training-dataset, method) cell we report the final-step validation accuracy on the training dataset's official test split (GSM8K, DeepScaleR-val, DeepCoder TACO-Verified). For out-of-distribution evaluation, we run the trained checkpoint zero-shot on a held-out math benchmark: GSM8K-trained checkpoints are evaluated on MATH-500~\citep{hendrycks2021math,lightman2023let}, while DeepScaleR-trained checkpoints are evaluated on AIME-24 and 25~\citep{maa_aime}.

\subsubsection{Main Results}
\label{subsubsec:rl-main}

To comprehensively compare GRPO and GRPO with SV, we show the in-domain validation accuracy for three different students in \cref{tab:rl-main}. We make the following observations:

\noindent SV-Explore is the strongest method on math, and SV-Rescue provides a smaller but positive lift over Plain GRPO in general. The gain is largest on Llama-3.2-3B, the weakest base model and therefore the student that produces the most wrong rollouts and all-wrong groups, giving the SV term the most rollouts to act on. On code, no method wins consistently across the three students, which is the same as the SFT regime for code, suggesting that code calls for a correctness direction different from the direction that works on math.

\subsubsection{Out-of-Distribution Evaluation}
\label{subsubsec:rl-ood}

\Cref{tab:rl-ood} reports the OOD accuracy of every trained checkpoint on its corresponding held-out benchmark, with MATH-500 evaluating the GSM8K-trained checkpoint, AIME-24 and AIME-25 evaluating the DeepScaleR-trained checkpoint, and LiveCodeBench evaluating the DeepCoder-trained checkpoint. AIME accuracy is averaged over $10$ independent samples per problem using temperature 0.7. The in-domain advantage of the SV-augmented methods carries over cleanly to the OOD regime. The largest reasoning OOD gain is on Llama-3.2-3B's GSM8K-to-MATH-500 transfer, where both SV variants clear Plain by a wide margin. Due to the page limit, we defer more analysis, including ablation of RL experiments in~Appendix \cref{sec:analysis on RL}.

\section{Conclusion}

In this paper, we present the first systematic investigation of steering vectors as persistent training signals rather than mere post-hoc inference controls. By integrating SVs into both supervised learning and RL pipelines, we demonstrate that internal representations can effectively improve post-training. In the supervised regime, our mechanism mitigates the off-policy gap inherent in teacher forcing, yielding consistent reasoning improvements, especially when distilling from thinking-mode teachers, and these gains hold over three seeds. In the RL regime, SV-Rescue recovers a learning signal on wrong rollouts, including all-wrong groups, and works best as a composable auxiliary signal on top of existing remedies and exploration schedules. While our comprehensive analysis maps specific boundary conditions, such as limited efficacy on code generation and the necessity of family-specific layer tuning, our results establish steering vectors as an effective training signal for LLM post-training.

\section*{Acknowledgment}
The work was partially supported by NSF awards \#2442477, \#2536297, \#2550203, \#2613637, \#2616632 and \#2623317. The views and conclusions in this paper should not be interpreted as representing any funding agencies.

\bibliographystyle{unsrtnat}
\bibliography{reference}

%%%%%%%%%%%%%%%%%%%%%%%%%%%%%%%%%%%%%%%%%%%%%%%%%%%%%%%%%%%%

\newpage
\newpage

\appendix

\section{Steering Vectors: Detailed Construction}
\label{sec:sv_appendix}
This appendix gives the complete construction of the steering-vector primitives used throughout the paper. The abbreviated version is in Section~\ref{subsec:sv} of the main text.       
\paragraph{Linear representation hypothesis and goal.}
A broad empirical observation in the interpretability literature is that many high-level behaviors of LLMs, such as sentiment, truthfulness, refusal, and task identity, are encoded along approximately linear directions in the residual stream~\citep{turner2023activation, li2023inference, panickssery2024caa, zou2023representation}. This motivates the goal of finding, at each layer $\ell$, a single direction $v^{(\ell)} \in \mathbb{R}^d$ such that shifting the hidden state along $v^{(\ell)}$ predictably shifts the corresponding behavior in the model's output without any parameter update. Building $v^{(\ell)}$ requires two ingredients: a way to summarize the hidden state produced on a response as one vector per layer, and a pair of response populations whose summaries contrast in the target behavior.

\paragraph{Answer-token pooling.}
A common choice for the per-response summary is the mean hidden state over the response tokens, masked to exclude the prompt. For a response of length $T$ with answer mask $m \in \{0,1\}^T$, the answer-token mean hidden state at layer $\ell$ is
\begin{equation}
\bar h^{(\ell)}(x, y) \;=\;
\frac{\sum_{t=1}^{T} m_t \, h^{(\ell)}(x, y_{<t})}{\sum_{t=1}^{T} m_t}
\;\in\; \mathbb{R}^d.
\label{eq:answer-mean-appendix}
\end{equation}
This collapses token-level variation into a single direction per layer and is the standard summary used in representation-level analyses of LLMs.

\paragraph{Difference-in-means steering vector.}
Given two matched response populations $\mathcal{D}_+$ and $\mathcal{D}_-$ that differ in the target behavior, the standard recipe~\citep{turner2023activation, panickssery2024caa, zou2023representation} sets $v^{(\ell)}$ to the difference of their answer-token-pooled means,
\begin{equation}
v^{(\ell)} \;=\;
\mathbb{E}_{i \in \mathcal{D}_+}\!\!\left[
\bar h^{(\ell)}(x_i, y_i^{+}) \right]
\;-\;
\mathbb{E}_{i \in \mathcal{D}_-}\!\!\left[
\bar h^{(\ell)}(x_i, y_i^{-}) \right].
\label{eq:sv-generic-appendix}
\end{equation}
We write $\hat v^{(\ell)} = v^{(\ell)} / \lVert v^{(\ell)} \rVert_2$ for the unit-normalized version. Under the linear representation assumption, hidden states in $\mathcal{D}_+$ carry a larger component along the behavior-relevant direction than those in $\mathcal{D}_-$, while they share the same distribution along unrelated axes. Averaging over many samples in each population washes out the behavior-unrelated variation, so the difference of the two means concentrates on the direction that discriminates the two populations. Shifting the residual stream along $v^{(\ell)}$ at inference time then targets the same direction that was already correlated with the desired behavior in the training-set activations.

\paragraph{Additive injection.}
The simplest way to apply $v^{(\ell)}$ to the hidden state at a chosen layer $\ell^\star$ is the additive primitive~\citep{turner2023activation, panickssery2024caa},
\begin{equation}
\tilde h^{(\ell^\star)}_t \;=\;
h^{(\ell^\star)}_t \;+\; \alpha \, v^{(\ell^\star)},
\qquad \alpha \in \mathbb{R}.
\label{eq:additive-inject-appendix}
\end{equation}
Additive injection changes both the direction and the magnitude of the hidden state, and has been reported to collapse the activation manifold at large $\alpha$~\citep{luo2026spherical}.

\paragraph{Spherical (Slerp) injection.}
Spherical linear interpolation (Slerp) was introduced for quaternion animation by~\citet{shoemake1985slerp} and later adopted for weight-space model merging~\citep{goddard2024mergekit} and for activation steering~\citep{luo2026spherical}. It avoids the magnitude-drift issue of additive injection by rotating the normalized hidden state $u_t = h^{(\ell^\star)}_t / \lVert h^{(\ell^\star)}_t \rVert_2$ toward a target direction $w$ along the shortest great-circle arc on the unit sphere,
\begin{equation}
\mathrm{Slerp}(u, w; \alpha) \;=\;
\frac{\sin\!\big((1-\alpha)\omega\big)}{\sin \omega}\, u
\;+\;
\frac{\sin(\alpha \omega)}{\sin \omega}\, w,
\qquad \omega = \arccos(u^{\!\top} w).
\label{eq:slerp-def-appendix}
\end{equation}
Taking $w \propto h^{(\ell^\star)}_t + v^{(\ell^\star)}$ gives the Slerp counterpart of Equation~\eqref{eq:additive-inject-appendix},
\begin{equation}
\tilde h^{(\ell^\star)}_t \;=\;
\lVert h^{(\ell^\star)}_t \rVert_2 \cdot
\mathrm{Slerp}\!\left(u_t,\, w;\, \alpha\right),
\label{eq:slerp-inject-appendix}
\end{equation}
which preserves the norm of $h^{(\ell^\star)}_t$ by construction.

\section{Hyperparameter Selection Protocol and Practical Guide}
\label{app:hparam}

This section expands the selection protocol summarized in \cref{subsubsec:sft-setting}.

\paragraph{What was fixed, selected, and analyzed post hoc.}
\begin{itemize}
  \item \textbf{Fixed in advance.} The injection layer $\ell^\star = L/2+1$ (mid-stack rule from inference-time steering~\citep{panickssery2024caa}) for every model, in both regimes. No reported number comes from a per-model layer search.
  \item \textbf{Selected on a development split.} The injection operation (Per-sample Slerp vs.\ Mean-SV) and $\alpha$ were selected once on Qwen2.5-1.5B with GSM8K, training on $6{,}000$ training examples and validating on the remaining training examples. This gives Per-sample Slerp at $\alpha=0.1$ and Mean-SV at $\alpha=0.5$, which are then reused for every cell of \cref{tab:sft-main,tab:distill-main}. For RL, $\lambda=0.1$ is used throughout. 
\end{itemize}

% \paragraph{The fixed default is conservative.} Because one configuration is reused everywhere, the per-cell optimum is often well above the reported number. For example, Llama-3.2-3B distilled from the Gemma thinking teacher on MATH already improves over Plain by $+4.8$pp under the default (\cref{tab:distill-main}); tuning this cell on a validation split carved from its training set raises the improvement above $+10$pp. Similarly, normalized SV with a tuned $\alpha$ reaches $+21.4$pp on the Qwen2.5-1.5B thinking-teacher MATH cell (\cref{sec:distill_analysis_appendix}). We report the fixed-default numbers.

\paragraph{Validation-based layer selection for a new backbone.} When adapting to a new architecture, we recommend a quarter-epoch ($0.25$) sweep over candidate layers, using the last $500$ training examples as the validation set and picking the best-validation layer. The total cost is roughly one extra epoch of training spread over the candidates. On Qwen2.5-1.5B this recovers $\ell^\star=15$, which is also the test optimum, and on Llama-3.2-3B it recovers $\ell^\star=10$, one of its two tied peaks, without access to test data. The same protocol applies to $\alpha$.

\paragraph{Practical heuristics.}
\begin{itemize}
  \item \textbf{Layer.} Start mid-stack; across both families we tested, effective layers lie in the middle of the network and neither shallow nor deep layers are ever optimal. Use the quarter-epoch sweep above if budget allows.
  \item \textbf{Operation and $\alpha$.} With no tuning budget, Mean-SV at $\alpha=0.5$ is the safer choice because of its broad $\alpha$ plateau. Per-sample Slerp has a higher peak at $\alpha=0.1$ but degrades quickly beyond $\alpha=0.3$, falling below Plain at $\alpha=0.5$ and collapsing at $\alpha=1$ (\cref{fig:sft-sensitivity:a}), so we advise keeping $\alpha \le 0.3$ for Slerp.
  \item \textbf{Large off-policy gaps.} With thinking-mode teachers, $\lVert v \rVert$ is roughly $3\times$ larger (\cref{fig:distill-stats}), and the raw-norm default under-injects; a unit-normalized direction with $\alpha$ around $1$--$3\times$ the raw norm is a good starting point.
  \item \textbf{When not to use the reference direction.} If the per-sample directions are dispersed (low mean $\cos(\delta_i, v)$, e.g.\ $0.48$ on MBPP versus $0.81$ on GSM8K), the reference direction likely encodes style rather than correctness; an outcome-conditioned direction (pass minus fail) is preferable.
\end{itemize}

% --------------------------------------------------------------------

\section{More Analysis on SFT Experiments}
\label{sec:sft_analysis_appendix}
Here, we provide more ablation studies and further analysis.

\noindent \textbf{MBPP redesign with execution-conditioned SV.} The reference-based direction is a poor target on MBPP for two reasons. First, the per-sample $\delta_i$ alignment is only $0.483$ on MBPP versus $0.808$ on GSM8K, so the direction is dispersed. Second, the cosine between the reference direction $v_{\mathrm{ref}}$ and the execution-conditioned direction $v_{\mathrm{exec}} = \mathbb{E}[h \mid \text{pass}] - \mathbb{E}[h \mid \text{fail}]$, which partitions the population by whether the bootstrap prediction passes the unit tests, is $-0.447$ at $\ell^\star=15$ on Qwen2.5-1.5B and $-0.437$ on Qwen2.5-7B. In other words, reference-SV actively pushes the model away from the test-passing direction, and $100\%$ of passing bootstrap predictions differ syntactically from the reference solution, so the reference target carries arbitrary style on top of correctness. Switching to $v_{\mathrm{exec}}$ at $\alpha=0.3$ on Qwen2.5-7B raises MBPP accuracy from $64.80\%$ (Plain SFT) to $66.00\%$, a $+1.20$pp gain that is the first non-noise positive on MBPP in the entire study, with $\alpha=0.1$ and $\alpha=0.5$ giving smaller positives ($65.20\%$ and $65.40\%$ respectively).

% CAMERA-READY: the LoRA absorption check is merged into the new train-vs-inference subsection below.

\subsection{Training-Time vs.\ Inference-Time SV}
\label{app:train-vs-inference}
\paragraph{Setup.} The inference-time SV is deliberately the \emph{same} direction used in training, namely the Mean-SV reference-minus-prediction direction at layer $\ell^\star = L/2+1$, added with coefficient $\alpha$. We do not claim this is a good inference-time steering vector; the purpose of the experiment is to separate the two uses of one direction. \Cref{tab:train-vs-inference} extends panel (c) of \cref{fig:sft-sensitivity} to six cells.

\begin{table}[h]
\centering
\caption{The same SV used during training versus at inference. ``SV at inference'' injects the training direction ($\alpha=0.5$) into the Plain-SFT checkpoint at test time. Training with the SV helps in every cell; injecting it at inference hurts in every cell.}
\label{tab:train-vs-inference}
\small
\begin{tabular}{lccc}
\toprule
Model, dataset & Plain & SV during training & SV at inference \\
\midrule
Qwen2.5-1.5B, GSM8K & 59.14 & 64.14 ($+5.00$) & 44.43 ($-14.71$) \\
Qwen2.5-1.5B, MATH  & 24.60 & 27.20 ($+2.60$) & 19.60 ($-5.00$) \\
Qwen2.5-7B, GSM8K   & 76.88 & 80.21 ($+3.33$) & 71.87 ($-5.01$) \\
Qwen3-4B, GSM8K     & 80.06 & 84.61 ($+4.55$) & 77.03 ($-3.03$) \\
Qwen2.5-1.5B, distill from Qwen3-8B, GSM8K & 65.81 & 67.55 ($+1.74$) & 63.53 ($-2.28$) \\
Qwen2.5-1.5B, distill from Qwen3-8B, MATH  & 21.20 & 31.20 ($+10.00$) & 10.80 ($-10.40$) \\
\bottomrule
\end{tabular}
\end{table}

\paragraph{Evidence for the offload account.} (i) \emph{Activation change.} With the SV active during training, the change that training makes to the layer-15 hidden state on training answer tokens has cosine $\approx -0.05$ with the SV direction, whereas SV-free training produces a clearly positive alignment. (ii) \emph{LoRA absorption check.} A natural worry is that the adapter simply learns to undo the injection. Plain and SV-trained LoRA deltas have identical Frobenius norm to within $1\%$ across all three students (1.5B $\frac{19.30}{19.29}=1.000$, 7B $\frac{27.85}{28.19}=0.988$, Qwen3-4B $\frac{35.38}{35.55}=0.995$), and the projection of the SV-trained LoRA delta onto $v$ is $\approx 10^{-6}$. The adapter therefore learns corrections overwhelmingly orthogonal to the SV, so the gain comes from redirected gradients rather than from magnitude inflation or from cancelling the injection. Together, these suggest that the SV supplies the component of the reference shift along its own direction (largely style and format) during training, so the parameters need not learn it; re-adding it at inference then only perturbs the hidden states.

\paragraph{A falsifiable prediction.} If the offload account is right, a configuration whose vector steers more strongly at inference supplies the target component more completely during the training forward pass, so offloading should also be more complete. Fixing the SV construction and varying the configuration (injection layer or strength), the inference-time steering effect of each configuration on the frozen model should therefore correlate positively with how much of the target behavior the corresponding trained model offloads, measured by the alignment of its learned update with the SV. We leave this causal test to future work.

\paragraph{Implication for placement.} The same account suggests why the best layer is model-specific. If the SV helps by offloading correctness-irrelevant information, the most effective injection layer is the one where the vector offloads the most such information. The best layer is then not purely a property of the architecture, since it also depends on how the training-time vector is constructed, since what can be offloaded is set by the vector itself.

\subsection{Full Variant Sweep on SV-Augmented SFT}
\label{sec:appendix-variants}

This appendix reports the full set of SV injection variants we tested on Qwen2.5-1.5B with GSM8K at $\ell^\star = 15$, in addition to the representative variants discussed in the main text. All variants below apply some form of steering vector throughout training; methods that disable the SV hook or use no SV at all are excluded by construction. The full numerical results are in \cref{tab:appendix-variants}; we group the variants by which of the three design axes from Section~\ref{subsec:method-sft} they perturb, namely direction, placement, or operation.

\paragraph{Direction-axis variants.}
The variants below modify how the injected direction $v^{\ell^\star}$ is constructed.
\begin{itemize}
  \item \textbf{Cos-weighted Mean-SV} reweights each sample's contribution to $v_{\mathrm{mean}}$ by $\cos(\delta_i, v_{\mathrm{mean}})$, so that samples whose per-sample direction is well aligned with the global mean count more.
  \item \textbf{Calc-wrong-only Mean-SV} builds $v_{\mathrm{mean}}$ from only the samples whose bootstrap prediction has a wrong final answer, on the hypothesis that wrong samples carry the most useful correction direction.
  \item \textbf{Clean-SV (consolidated)} keeps only the samples with $\cos(\delta_i, v_{\mathrm{mean}}) > 0.3$ and recomputes the mean from this clean subset, then injects the consolidated direction on every training sample.
  \item \textbf{PCA PC1 SV} (also reported in the main text) replaces $v_{\mathrm{mean}}$ with the first principal component of the per-sample $\delta_i$ matrix.
  \item \textbf{Dynamic SV} recomputes the steering vector online during training. Every $200$ optimizer steps we re-run the training set through the current LoRA-adapted model with and without GT teacher forcing, recompute both the GT-mean and the Pred-mean, and replace the injected direction with the freshly recomputed $v_{\mathrm{mean}}$.
\end{itemize}

\paragraph{Placement-axis variants.}
The variants below modify where the SV is injected, namely which token positions or which layers.
\begin{itemize}
  \item \textbf{Slerp first-$K$} restricts the Slerp hook to only the first $K$ answer tokens, varying the front-loaded scope; we tested $K = 32$ and $K = 64$.
  \item \textbf{Confidence-gated SV} applies the SV only to answer-token positions where the next-token entropy under the current model exceeds a threshold, on the hypothesis that low-confidence tokens are where correction matters.
  \item \textbf{Multi-layer (uniform)} (also reported in the main text) injects Mean-SV at $L \in \{10, 15, 20\}$ with uniform per-layer $\alpha = 0.50/0.50/0.50$.
  \item \textbf{Multi-layer (tuned)} injects at the same three layers with $\alpha = 0.10/0.50/0.10$, putting more weight on the middle layer.
  \item \textbf{All-layer injection} injects at every transformer block with the single-layer coefficient $\alpha = 0.5$. Training collapses, and the model emits an empty or truncated final answer on about $92\%$ of GSM8K test problems and scores near zero. We attribute this to the total intervention strength, since reusing a single-layer coefficient at every layer multiplies the effective perturbation, rather than to multi-layer injection itself. The three-layer schedules above train stably, but the directions at layers $10$, $15$, and $20$ have pairwise cosine $\approx 0.78$, so multi-layer injection largely amplifies one shared direction rather than adding independent corrections. Single-layer injection is therefore the simpler and more robust default. 
\end{itemize}

\paragraph{Operation-axis variants.}
The variants below modify how the SV combines with the existing hidden state, including the injection primitive, the sample-level gating, and the $\alpha$ schedule.
\begin{itemize}
  \item \textbf{Mean-SV additive} is the standard mean-direction baseline that uses the global $v_{\mathrm{mean}}$ with the additive primitive of Equation~\eqref{eq:additive-inject}.
  \item \textbf{Per-sample additive} (also reported in the main text) keeps the per-sample $\delta_i$ but injects it through the additive primitive rather than through Slerp.
  \item \textbf{Negated Mean-SV additive} flips the sign of the steering vector, subtracting $\alpha v_{\mathrm{mean}}$ instead of adding, to test whether the gain depends on the direction of the SV rather than on the presence of any in-place perturbation.
  \item \textbf{Gated Slerp} drops samples whose $\cos(\delta_i, v_{\mathrm{mean}})$ falls below a threshold and applies SV only on the remaining samples; we tested $\tau = 0.5$ and the more aggressive $\tau = 0.7$.
  \item \textbf{$\alpha$ decay schedule} starts at $\alpha = 0.5$ and linearly decays to $\alpha = 0$ over the epoch, asking whether SV is more useful in early training.
\end{itemize}

\begin{table}[t]
\centering
\caption{Full sweep of SV injection variants on Qwen2.5-1.5B with GSM8K at $\ell^\star = 15$. Methods are sorted by accuracy.}
\label{tab:appendix-variants}
\small
\begin{tabular}{lccc}
\toprule
Method & $\alpha$ & Accuracy & $\Delta$ \\
\midrule
Plain LoRA (no SV)                                 & --             & 59.14 & 0 \\
Negated Mean-SV additive (sign-flip control)       & $-0.50$        & 54.58 & $-4.56$ \\
Random per-sample Gaussian (control)               & 0.50           & 54.82 & $-4.32$ \\
$\alpha$ decay schedule (linear $0.5 \to 0$)       & $0.5 \to 0$    & 58.07 & $-1.07$ \\
Gated Slerp (drop $\cos < 0.7$)                    & 0.10           & 58.45 & $-0.69$ \\
Slerp first-$K{=}64$ (extended scope)              & 0.10           & 58.61 & $-0.53$ \\
Clean-SV (consolidated mean, $\cos > 0.3$ subset)  & 0.10           & 58.76 & $-0.38$ \\
Confidence-gated SV (mask high-conf tokens)        & 0.50           & 58.98 & $-0.16$ \\
PCA PC1 SV                                         & 0.50           & 59.29 & $+0.15$ \\
Gated Slerp (drop $\cos < 0.5$)                    & 0.10           & 59.36 & $+0.22$ \\
Dynamic SV (online recomputation every 200 steps)  & 0.50           & 60.23 & $+1.09$ \\
Multi-layer (L=10/15/20, uniform)                  & 0.50/0.50/0.50 & 60.85 & $+1.71$ \\
Mean-SV additive                                   & 0.50           & 61.11 & $+1.97$ \\
Slerp first-$K{=}32$ (front-loaded)                & 0.10           & 62.02 & $+2.88$ \\
Cos-weighted Mean-SV additive                      & 0.50           & 62.40 & $+3.26$ \\
Calc-wrong-only Mean-SV additive                   & 0.50           & 62.55 & $+3.41$ \\
Multi-layer (L=10/15/20, tuned)                    & 0.10/0.50/0.10 & 62.62 & $+3.48$ \\
Per-sample additive                                & 0.50           & 62.93 & $+3.79$ \\
\textbf{Per-sample Slerp (full answer)}            & \textbf{0.10}  & \textbf{64.14} & $\mathbf{+5.00}$ \\
\bottomrule
\end{tabular}
\end{table}

\paragraph{Discussion of the full sweep.}
The full sweep reinforces the four conclusions drawn in the main text and adds three further ones. None of the direction-construction variants we tested exceeds plain Mean-SV by a meaningful margin, which we read as a consequence of the per-sample $\delta_i$'s being already $99.9\%$ aligned with the global mean on GSM8K (mean $\cos = 0.808$); when the per-sample directions all point the same way, there is essentially nothing for a filter or a cleaner aggregation to recover. Sample-level and token-level gating (Gated Slerp at both thresholds, Confidence-gated SV) all under-perform the no-gating baseline, which we read as evidence that selectively skipping samples removes useful gradient signal more than it removes noise. Scope restrictions (Slerp first-$K$ at $K=32$ or $K=64$) hurt at every $K$ we tried, confirming that the SV signal is broadly distributed across the answer span rather than concentrated at the prefix. The tuned multi-layer schedule outperforms the uniform multi-layer schedule by a substantial margin but still falls short of single-layer Per-sample Slerp, which is consistent with the main-text reading that multi-layer injection demands careful per-layer $\alpha$ tuning rather than reusing a shared $\alpha$ across layers. Finally, the negated Mean-SV additive control matches the random-Gaussian control at roughly $5$pp below Plain, ruling out the alternative that the SV gain comes from any in-place hidden-state perturbation rather than from the specific GT-vs-Pred direction.

\section{More Analysis on Distillation Experiments}
\label{sec:distill_analysis_appendix}

This appendix collects the additional distillation analyses we deferred from the main text. All results are on the same $3 \times 4 \times 4$ grid reported in \cref{tab:distill-main}.

\begin{figure}[t]
\centering
\begin{subfigure}[t]{0.56\linewidth}
  \centering
  \includegraphics[width=\linewidth]{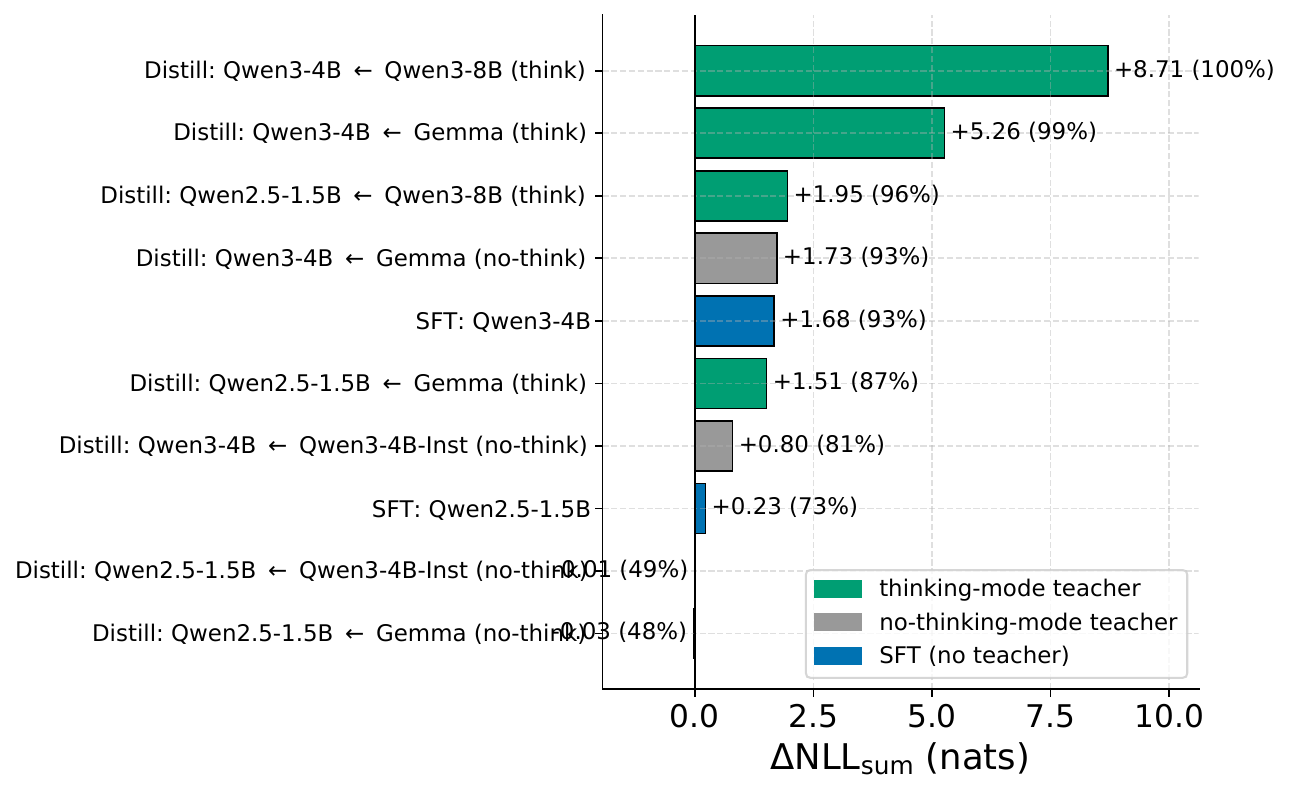}
  \caption{$\Delta\mathrm{NLL}_{\mathrm{sum}}$ across all 10 cells, sorted descending. Numbers in parentheses are $\mathrm{frac}(\Delta\mathrm{NLL}_{\mathrm{sum}} > 0)$.}
  \label{fig:distill-offpolicy-full:a}
\end{subfigure}\hfill
\begin{subfigure}[t]{0.42\linewidth}
  \centering
  \includegraphics[width=\linewidth]{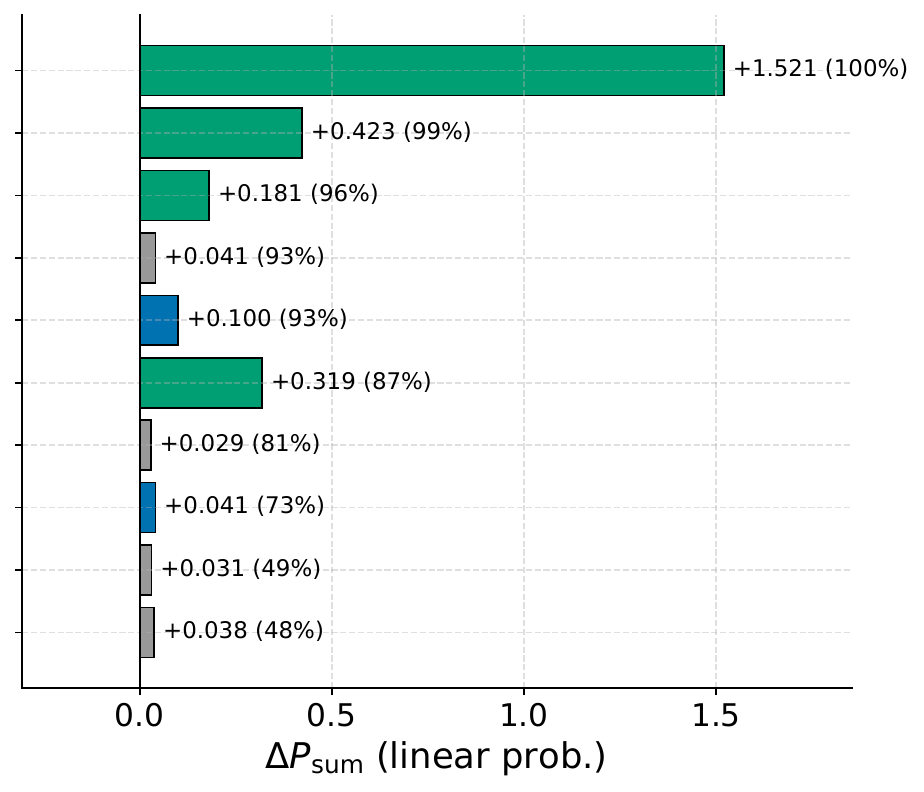}
  \caption{$\Delta P_{\mathrm{sum}}$ on the same 10 cells in the same row order as panel (a), in linear probability units.}
  \label{fig:distill-offpolicy-full:b}
\end{subfigure}
\caption{Full off-policy mitigation sweep on all $10$ cells (eight distillation cells of the headline grid plus the two SFT cells), in two complementary scales. Eight of ten cells are positive at $\mathrm{frac}(\Delta\mathrm{NLL}_{\mathrm{sum}} > 0) \ge 73\%$; the two near-zero/negative cells in panel (a) are exactly the Qwen2.5-1.5B $\times$ non-thinking-teacher distillation cells that also show no SV-distill accuracy gain in \cref{tab:distill-main}. The linear-probability metric in panel (b) is non-negative on every cell because the per-token probability mass moved onto label tokens is positive even on the cells where token-level NLL gains and losses cancel.}
\label{fig:distill-offpolicy-full}
\end{figure}

\paragraph{Full off-policy mitigation sweep across all 10 cells.}
The main text reports the four Qwen3-4B distillation cells of \cref{fig:distill-analysis:a,fig:distill-analysis:b}. \Cref{fig:distill-offpolicy-full} extends the test to all $8$ distillation cells of the headline grid (two students $\times$ four teachers) plus the two SFT cells, in both the log-probability metric $\Delta\mathrm{NLL}_{\mathrm{sum}}$ and the linear-probability metric $\Delta P_{\mathrm{sum}}$. The protocol matches the main text: we apply the trained-time SV hook (additive, $\alpha=0.5$, $\ell^\star=L/2{+}1$) to the base model under teacher forcing on $500$ sampled training prompts and report the per-prompt sums and the fraction of prompts on which $\Delta\mathrm{NLL}_{\mathrm{sum}}$ is positive. The two cells without off-policy mitigation (Qwen2.5-1.5B distilled from a non-thinking teacher) are exactly the cells where the headline accuracy table shows no SV-distill gain.

\begin{figure}[t]
\centering
\begin{subfigure}[t]{0.32\linewidth}
  \centering
  \includegraphics[width=\linewidth]{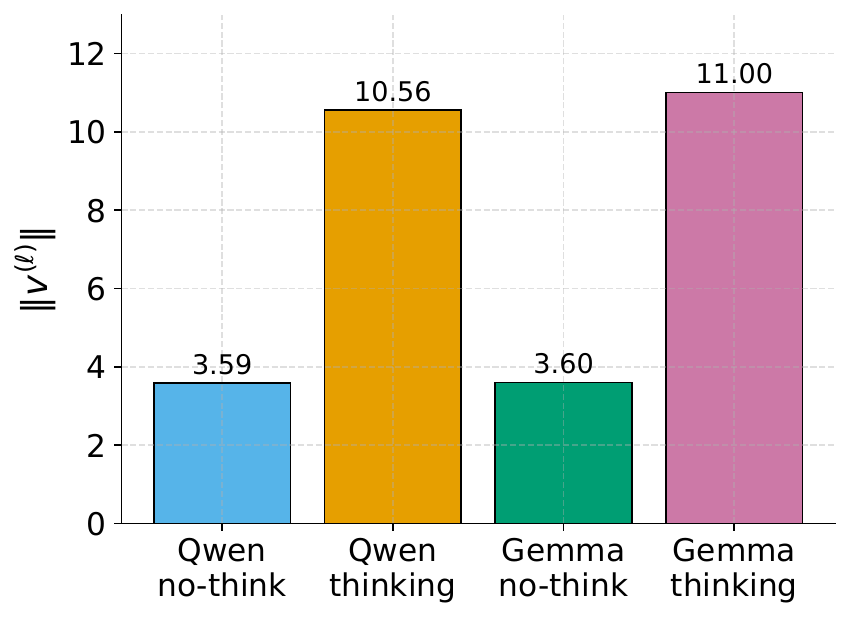}
  \caption{Direction norm $\|v^{(\ell)}\|$ at $\ell^\star=15$, averaged over the 12 (student, dataset) cells of each teacher.}
  \label{fig:distill-stats:a}
\end{subfigure}\hfill
\begin{subfigure}[t]{0.32\linewidth}
  \centering
  \includegraphics[width=\linewidth]{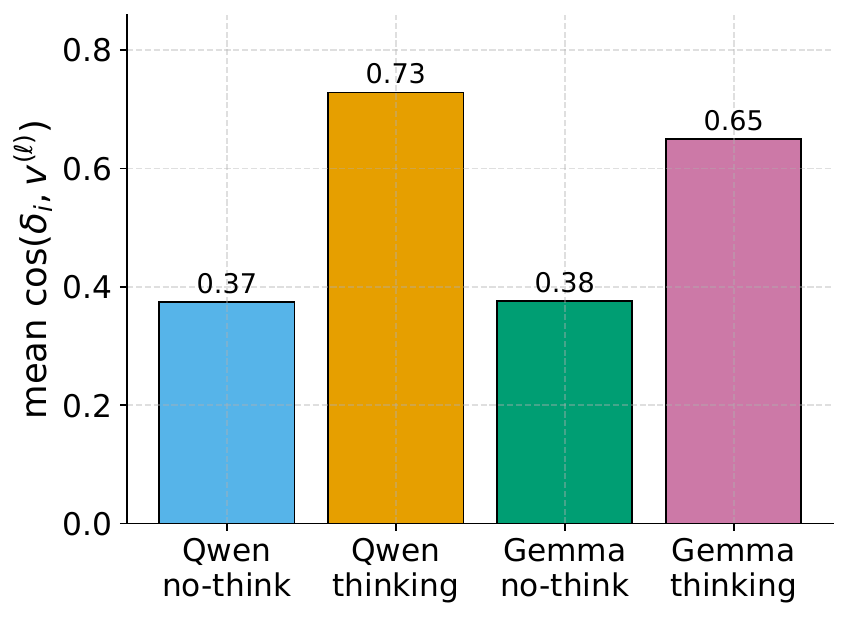}
  \caption{Mean per-sample alignment $\cos(\delta_i, v^{(\ell)})$ at $\ell^\star=15$, averaged over the 12 cells of each teacher.}
  \label{fig:distill-stats:b}
\end{subfigure}\hfill
\begin{subfigure}[t]{0.32\linewidth}
  \centering
  \includegraphics[width=\linewidth]{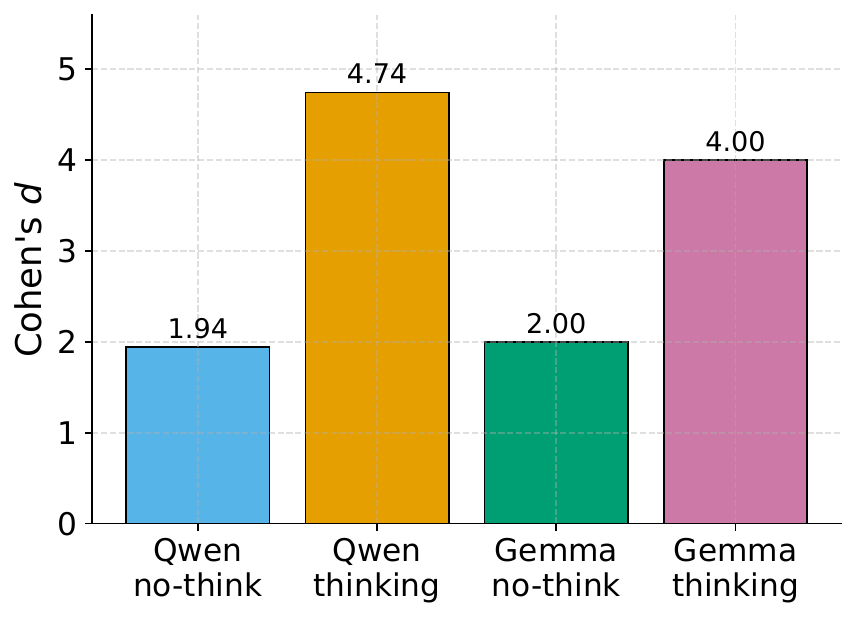}
  \caption{Cohen's $d$ for the projection of the GT and Pred populations onto $v^{(\ell)}$ at $\ell^\star=15$, averaged over the 12 cells of each teacher.}
  \label{fig:distill-stats:c}
\end{subfigure}
\caption{Three statistics of the SV direction at $\ell^\star=15$ broken down by teacher mode and averaged over the 12 (student, dataset) cells of each teacher, namely the direction norm in panel (a), the per-sample alignment in panel (b), and Cohen's $d$ for the GT-versus-Pred projection in panel (c). Both thinking teachers dominate the two non-thinking teachers on every axis, while the two non-thinking teachers (different families) are statistically indistinguishable on all three quantities.}
\label{fig:distill-stats}
\end{figure}

\paragraph{SV direction statistics by teacher.}
\Cref{fig:distill-stats} reports three statistics for $v^{(\ell)}$ at $\ell^\star=15$ averaged over the $12$ cells of each teacher, namely the norm $\|v\|$, the mean alignment $\cos(\delta_i, v)$, and Cohen's $d$ for the $\mathrm{proj}_{\mathrm{GT}} - \mathrm{proj}_{\mathrm{Pred}}$ split. Both thinking teachers (Qwen3 thinking and Gemma thinking) produce roughly $3\times$ the norm, $\sim$$2\times$ the alignment, and $\sim$$2\times$ the Cohen's $d$ of either non-thinking teacher, and the two non-thinking teachers are statistically indistinguishable on all three quantities despite being from different families. The two thinking teachers, by contrast, agree in magnitude on these statistics but produce different downstream gains, which is consistent with the cosine-similarity finding in the main text that their directions are not interchangeable.

\begin{figure}[t]
\centering
\begin{subfigure}[t]{0.46\linewidth}
  \centering
  \includegraphics[width=\linewidth]{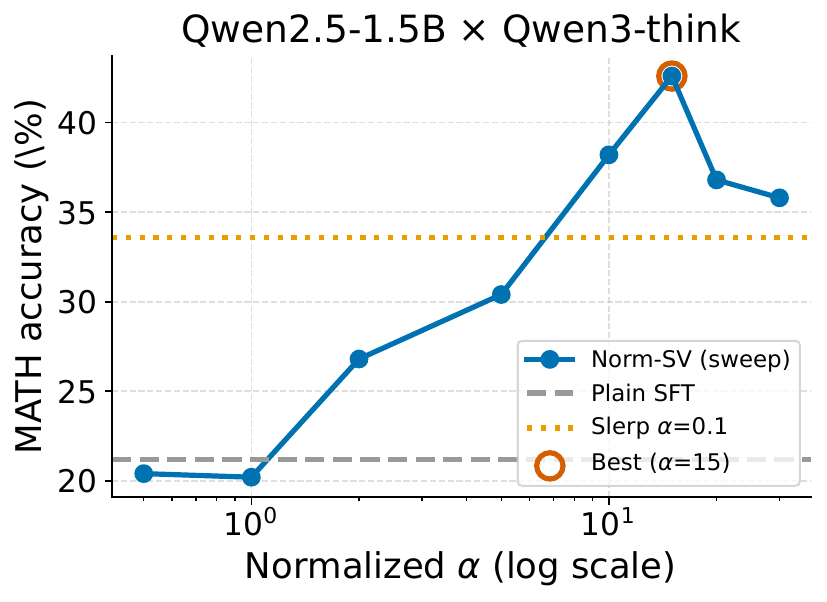}
  \caption{Normalized Mean-SV $\alpha$-sweep on Qwen2.5-1.5B distilled from the Qwen3 thinking teacher on MATH.}
  \label{fig:distill-norm-sv:a}
\end{subfigure}\hfill
\begin{subfigure}[t]{0.46\linewidth}
  \centering
  \includegraphics[width=\linewidth]{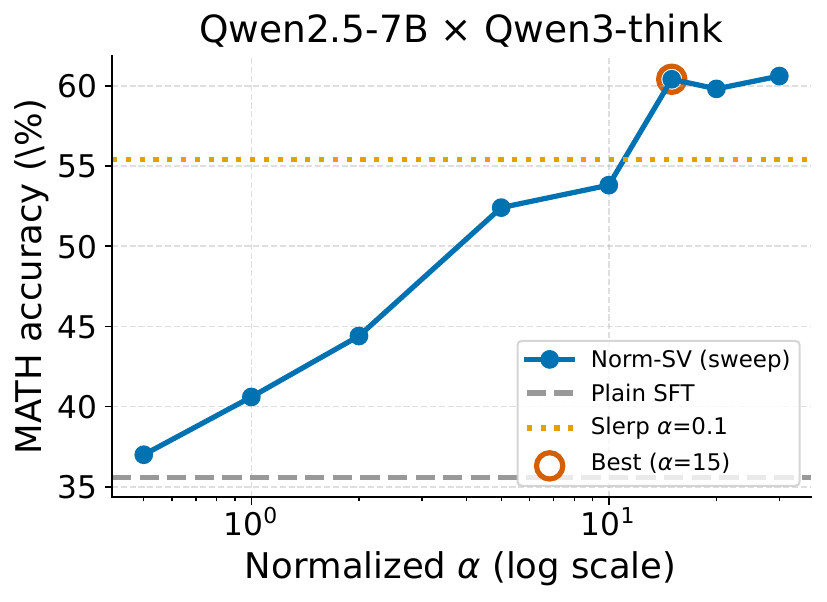}
  \caption{Normalized Mean-SV $\alpha$-sweep on Qwen2.5-7B distilled from the Qwen3 thinking teacher on MATH.}
  \label{fig:distill-norm-sv:b}
\end{subfigure}
\caption{Accuracy under a normalized Mean-SV $\alpha$-sweep on the Qwen3-thinking-teacher MATH cells, where the steering direction is unit-normalized so that $\alpha$ is interpretable as the literal injection magnitude rather than a Slerp rotation fraction. Plain SFT is shown as a dashed grey line and the Per-sample Slerp $\alpha = 0.1$ result from \cref{tab:distill-main} as a dotted orange line. Both panels show that a tuned $\alpha \in [10, 30]$ on the normalized direction outperforms Per-sample Slerp at $\alpha = 0.1$.}
\label{fig:distill-norm-sv}
\end{figure}

\paragraph{Normalized SV vs.\ unnormalized SV.}
Both Mean-SV and Per-sample Slerp inject the raw direction $v$ at a fixed coefficient $\alpha$, so the actual injection magnitude is $\alpha \cdot \|v\|$ and confounds two factors, namely how strongly we steer and how large the underlying direction happens to be. Because thinking teachers produce SV directions with substantially larger $\|v\|$ than non-thinking teachers (\cref{fig:distill-stats:a}), the same nominal $\alpha$ is not comparable across teachers. We therefore try \emph{Normalized SV}, which replaces $v$ with the unit-normalized $\hat v = v / \|v\|$, so that $\alpha$ becomes the injection magnitude in unit-norm space. On the Qwen3-thinking $\times$ MATH cell that produced the largest gain in \cref{tab:distill-main}, Normalized SV at a tuned $\alpha=15$ reaches a $+21.40$pp gain over Plain on Qwen2.5-1.5B, up from $+12.40$pp under Per-sample Slerp at $\alpha=0.1$. The full sweep across both Qwen2.5-1.5B and Qwen2.5-7B is shown in \cref{fig:distill-norm-sv}: the $\alpha$-curve is monotonic and concave on both students, with the optimum in the range $\alpha \in [10, 30]$, which corresponds to roughly $1$ to $3$ times the raw norm $\|v_{\mathrm{raw}}\|$. Normalized SV at this tuned $\alpha$ beats Per-sample Slerp at $\alpha = 0.1$ on every think-teacher MATH cell we tested, so the headline number is bottlenecked by under-injection of the raw direction rather than by the direction itself.

\paragraph{Cluster separation as a predictor of SV gain.}
The story from \cref{subsubsec:sft-analysis} that ``higher Cohen's $d$ implies more SV gain'' breaks down on distillation. The three cells with the highest Cohen's $d$ in the distillation grid are all Qwen3-thinking $\times$ MBPP cells, with $d \in \{5.85, 6.77, 7.10\}$, but the SV gain on those same cells is in $\{-1.80, 0.00, -0.60\}$. \Cref{tab:cohens-vs-gain} reports the top-six cells by Cohen's $d$ alongside their SV gain. A high $d$ separates GT from Pred populations well, but only on the training distribution. For the gain to transfer to test time, the direction must encode signal that survives on-policy decoding, not just the GT-vs-Pred gap on training prompts. On thinking-teacher MBPP, the dominant component of the GT-Pred gap is verbose reasoning prefix versus terse code, which is a surface artifact rather than a transferable correctness signal.

\begin{table}[t]
\centering
\caption{The six (student, teacher, dataset) cells with the largest Cohen's $d$ in the distillation grid, alongside the direction norm $\|v\|$, the per-sample alignment $\cos$, and the Per-sample Slerp gain over Plain. A high Cohen's $d$ on the training distribution does not imply a positive gain at test time, as illustrated by the three top-ranked MBPP cells whose gain is at most zero. The cell with the largest gain in this subset is shown in \textbf{bold}.}
\label{tab:cohens-vs-gain}
\small
\begin{tabular}{lcccc}
\toprule
Cell (student $\times$ teacher $\times$ dataset) & $\|v\|$ & mean cos & Cohen's $d$ & SV $\Delta$ (pp) \\
\midrule
7B $\times$ Qwen3-thinking $\times$ MBPP        & 22.39 & 0.801 & 7.10 & $-1.80$ \\
Qwen3-4B $\times$ Qwen3-thinking $\times$ MBPP  & 12.93 & 0.814 & 6.77 & $0.00$  \\
1.5B $\times$ Qwen3-thinking $\times$ MBPP      & 18.33 & 0.835 & 5.85 & $-0.60$ \\
Qwen3-4B $\times$ Qwen3-thinking $\times$ GSM8K & 5.51  & 0.783 & 5.84 & $-2.20$ \\
7B $\times$ Qwen3-thinking $\times$ GSM8K       & 10.57 & 0.778 & 5.79 & $+3.94$ \\
7B $\times$ Qwen3-thinking $\times$ MATH & 11.23 & 0.749 & 4.76 & $\mathbf{+19.80}$ \\
\bottomrule
\end{tabular}
\end{table}

\paragraph{Why Llama-3.2-3B underperforms under the canonical recipe.}
The cross-family Llama-3.2-3B student gives essentially neutral mean Per-sample Slerp $\Delta$ across teachers in \cref{tab:distill-main}, in contrast to the consistently positive Qwen students. This is the same layer-choice issue that the SFT layer sweep in panel (b) of \cref{fig:sft-sensitivity} already exposes, and not a failure specific to distillation. The Qwen-default rule sets $\ell^\star = L/2+1$ on every Qwen student and lands on a representation regime where the GT-Pred direction is dense and well-aligned, but the same fractional depth on Llama-3.2-3B's $28$ blocks ($\ell^\star = 15$) sits in a valley between two off-center optima. On Llama with GSM8K, Plain SFT is $62.09\%$, the canonical $\ell^\star=15$ drops the model by $-0.30$pp to $61.79\%$, while $\ell^\star=10$ and $\ell^\star=19$ both lift Llama by $+1.22$pp to $63.31\%$. A tuned distill layer of $\ell^\star=13$ on the Gemma-thinking teacher reaches $+9.0$pp over the canonical $\ell^\star=15$ on the same cell. The optimal SV injection layer therefore depends jointly on student family, dataset, and teacher mode, and is not a fixed fraction of network depth. Reusing a single-layer index across architectures is a confound that systematically suppresses Llama's true potential rather than evidence that SV transfer fails. 

\section{Analysis on Reinforcement Learning Experiments}
\label{sec:analysis on RL}

This appendix collects the additional RL analyses that we deferred from the main text, namely two ablations that isolate the contribution of the SV direction and of the SV-Rescue advantage modification, and a layer-wise correlation study that supports the canonical injection layer.

\paragraph{Ablation 1: SV direction vs random direction.}
We replace the precomputed correctness direction in SV-Rescue with a same-norm random Gaussian vector, holding every other hyperparameter fixed. \Cref{tab:rl-ablation-random} shows that the precomputed correctness direction beats the random control on both students. The SV-Rescue gain therefore depends on the direction being correctness-aligned and is not an artifact of shifting the advantage of wrong rollouts per se.

\begin{table}[t]
\centering
\caption{SV-Rescue with the precomputed correctness direction (ours) vs a same-norm random Gaussian direction on GSM8K. The SV direction is what carries the rescue gain.}
\label{tab:rl-ablation-random}
\small
\begin{tabular}{lcc}
\toprule
Student & Random direction & SV-Rescue (ours) \\
\midrule
Qwen2.5-1.5B & 77.54 & \textbf{78.35} \\
Qwen3-4B     & 90.35 & \textbf{91.04} \\
\bottomrule
\end{tabular}
\end{table}

\paragraph{Ablation 2: SV-Explore without SV-Rescue.}
SV-Explore stacks SV-Rescue on top of a per-prompt temperature schedule. To check that SV-Rescue is not redundant once exploration is added, we run SV-Explore on GSM8K with the SV-Rescue advantage disabled (temperature schedule only). \Cref{tab:rl-ablation-rescue} shows that removing SV-Rescue drops accuracy on both students, so the headline SV-Explore gain in \cref{tab:rl-main} is the joint contribution of SV-Rescue and the exploration schedule rather than the schedule alone.

\begin{table}[t]
\centering
\caption{SV-Explore (full) vs SV-Explore with the SV-Rescue advantage disabled (temperature schedule only) on GSM8K. SV-Rescue is not redundant once exploration is added.}
\label{tab:rl-ablation-rescue}
\small
\begin{tabular}{lcc}
\toprule
Student & Temperature schedule only & SV-Explore (full) \\
\midrule
Qwen2.5-1.5B & 78.72 & \textbf{79.18} \\
Qwen3-4B     & 91.79 & \textbf{91.99} \\
\bottomrule
\end{tabular}
\end{table}

\paragraph{Combination with entropy-aware shaping.} Entropy-aware advantage shaping~\citep{lai2025noprompt} also assigns learning signal to zero-variance groups. Over three seeds on Qwen2.5-1.5B with GSM8K, GRPO reaches $77.60 \pm 0.51$, GRPO with SV-Rescue $78.13 \pm 0.93$, GRPO with entropy-aware shaping $78.64 \pm 0.80$, and GRPO with both $78.87 \pm 0.59$. Entropy-aware shaping alone is stronger than SV-Rescue alone, while stacking SV-Rescue on top of it gives the best result. SV-Rescue is therefore best viewed as a composable auxiliary signal rather than a standalone solution to the zero-variance problem.

\begin{figure}[t]
\centering
\includegraphics[width=0.92\linewidth]{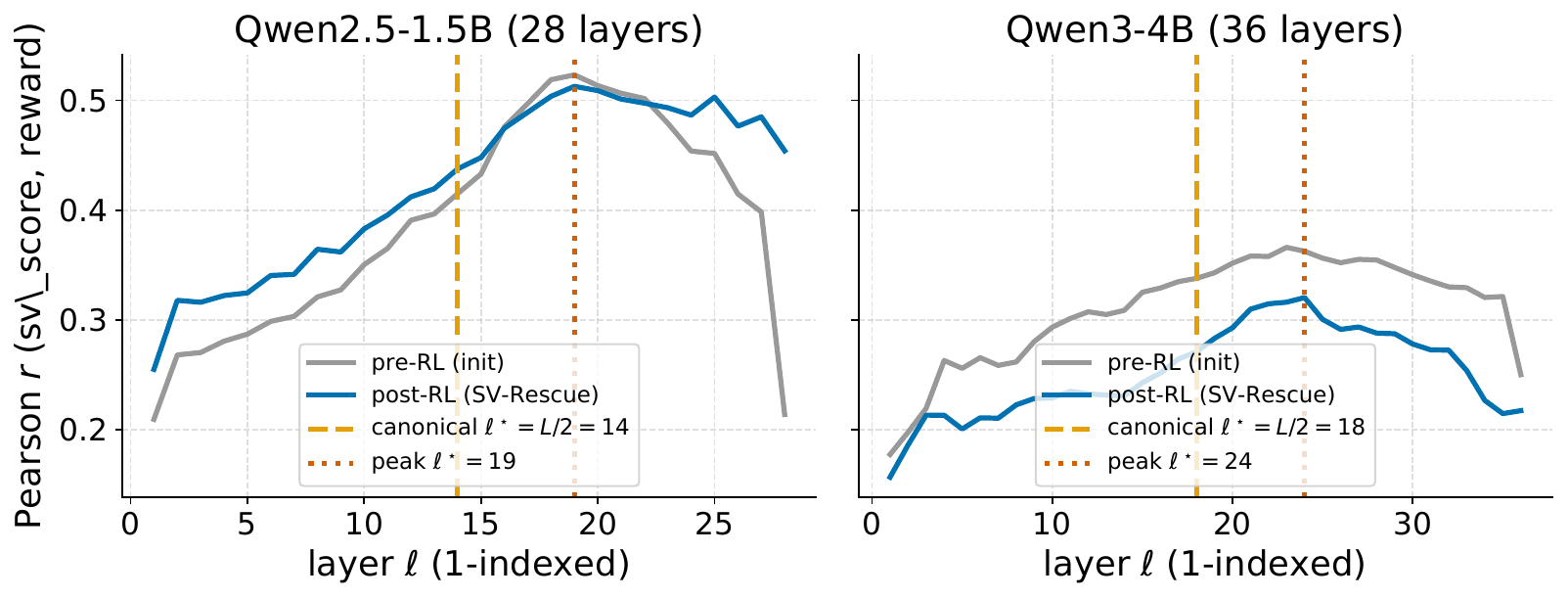}
\caption{Per-rollout Pearson correlation between the SV alignment score at each transformer block and the binary verifier reward, on a $2000$-rollout GSM8K hold-out, both before RL (init) and after RL (SV-Rescue). The vertical dashed line marks the canonical injection layer $\ell^\star=L/2+1$ used by SV-Rescue and the dotted line marks the peak-correlation layer. The correlation peaks slightly above $L/2+1$ on both students, but the canonical $L/2+1$ retains $80$-$90\%$ of the peak.}
\label{fig:rl-layer-correlation}
\end{figure}

\paragraph{Layer-wise SV signal.}
SV-Rescue uses $\ell^\star = L/2+1$ as the layer at which the SV alignment score is read.  To check whether this is the best layer for the RL setting, we compute the per-rollout Pearson correlation between the SV alignment score at every transformer block and the binary verifier reward on a $2000$-rollout GSM8K hold-out, both before and after RL (\cref{fig:rl-layer-correlation}). The peak correlation lies slightly above $\ell^\star=L/2+1$, near $2/3$ depth, although the canonical $L/2+1$ retains $80$-$90\%$ of the peak. To test whether the higher hold-out correlation translates into stronger RL gain, we re-trained Qwen2.5-1.5B GSM8K SV-Rescue at the peak-correlation layer; the peak-layer run reaches $78.66\%$, marginally above the canonical baseline of $78.35\%$. Layer choice is therefore a small additional lever rather than a major one, and the canonical $L/2+1$ default works well across students.

\section{Additional Baselines}
\label{app:baselines}

All baselines below use the same model (Qwen2.5-1.5B), data, LoRA configuration, one-epoch budget, and evaluation protocol as our method.

\paragraph{Supervised baselines.} \Cref{tab:sft-baselines} compares the SV with two baselines that also target the off-policy mismatch of SFT. DFT~\citep{wu2025dft} acts at the level of the objective, whereas the SV acts on the representation. With its original hyperparameters, DFT barely changes Plain under our LoRA setting; re-tuned for LoRA, it reaches $61.09\%$, still below the SV. Adding the SV on top of DFT yields $+3.76$pp over DFT, so the representation-level correction is not subsumed by the objective-level one. A per-token hidden-state matching loss instead hurts accuracy, so explicitly pulling hidden states toward the reference does not replace injecting the direction.

\begin{table}[h]
\centering
\caption{SFT-side baselines on Qwen2.5-1.5B with GSM8K (mean $\pm$ std over three seeds unless marked $^\dagger$).}
\label{tab:sft-baselines}
\small
\begin{tabular}{lcc}
\toprule
Method & Accuracy & $\Delta$ vs.\ Plain \\
\midrule
Plain SFT                      & $58.81 \pm 0.37$ & $0$ \\
DFT (original hyperparameters) & $58.88 \pm 0.80$ & $+0.07$ \\
DFT (re-tuned for LoRA)        & $61.09^\dagger$  & $+2.28$ \\
Hidden-state distillation      & $53.68^\dagger$  & $-5.13$ \\
SV (Per-sample Slerp)          & $\mathbf{62.75 \pm 0.50}$ & $\mathbf{+3.94}$ \\
DFT (original) + SV            & $62.64 \pm 0.73$ & $+3.83$ \\
\bottomrule
\end{tabular}
\end{table}

\paragraph{DFT~\citep{wu2025dft}.} DFT rescales each token's cross-entropy loss by the token's detached probability under the current model, which corrects the off-policy mismatch at the level of the objective. We first used the hyperparameters recommended in the original paper, which were developed for full-parameter fine-tuning on different models and datasets and transfer poorly to our LoRA setting ($58.88\%$). We then re-searched DFT's hyperparameters under our setting, reaching $61.09\%$. \emph{DFT + SV} uses DFT with its original hyperparameters, applying our Per-sample Slerp hook in the forward pass and the DFT loss in the objective.

\paragraph{Hidden-state distillation.} This baseline asks whether explicitly pulling hidden states toward the reference with an auxiliary loss can replace injecting the direction; it reduces accuracy to $53.68\%$.

\paragraph{Entropy-aware advantage shaping~\citep{lai2025noprompt}.} This method assigns token-level advantages to zero-variance prompts based on token entropy, so all-wrong groups also contribute gradient. \emph{Entropy-aware + SV-Rescue} adds the SV-Rescue term of \cref{eq:rl-sv-rescue} on top of the entropy-aware advantages. 

\paragraph{Relation of the three-seed RL runs to \cref{tab:rl-main}.} The three-seed GRPO runs are an independent rerun with the same configuration; their means ($77.60$ for GRPO and $78.13$ for SV-Rescue) differ slightly from the single-seed values in \cref{tab:rl-main} ($77.72$ and $78.35$).
% --------------------------------------------------------------------

\section{Discussion}
\label{sec:discussion}

Our results reveal a consistent set of conditions under which steering vectors provide a useful training signal, as well as clear boundaries where they do not.

\paragraph{When does SV injection help?}
The largest and most reliable gains appear on reasoning tasks (GSM8K, MATH) where the off-policy gap between the model's native style and the reference completion is large and directionally coherent. Per-sample Slerp exploits per-instance heterogeneity in this gap and outperforms the global Mean-SV direction on most cells, while Mean-SV remains competitive on the rest. In distillation, thinking-mode teachers amplify the benefit because their long reasoning traces widen the stylistic contrast and give the SV a richer correction signal. In RL, SV-Rescue recovers gradient on all-wrong groups that would otherwise contribute nothing to learning, and composes cleanly with orthogonal rollout-side mechanisms and with entropy-aware advantage shaping. Its standalone gain is small and below entropy-aware shaping alone, so we view it as an auxiliary signal rather than a standalone remedy.

\paragraph{When does SV injection fail?}
The two failure modes we observe line up with two of the design axes from Section~\ref{subsec:method-sft}. First, code generation breaks the \emph{direction}, where the reference-based SV direction is anti-correlated with the execution-conditioned correctness direction, so the SV actively pushes the model away from test-passing behavior. Second, cross-family layer transfer breaks the \emph{placement}, where the canonical $\ell^\star = L/2+1$ that works for Qwen lands in a valley for Llama-3.2-3B and requires a per-architecture layer search.

\paragraph{Training-time SV is internalized rather than memorized.}
A separate observation that does not fit cleanly into the three-axis decomposition is that a model trained with Per-sample Slerp performs best at inference \emph{without} SV, even though it was trained with SV active at every answer-token position. This indicates that the SV does not act as a permanent crutch the model leans on at decode time. Instead, the SV redirects the gradient during training so that the corrected behavior is internalized into the parameters themselves, and the resulting checkpoint generalizes beyond the steered training distribution rather than being tied to it.  \Cref{subsubsec:sft-analysis} and \cref{app:train-vs-inference} analyze this shift and show that the learned update is nearly orthogonal to the SV, which is consistent with the SV offloading a correctness-irrelevant component during training. A causal test of this offload account is the main open question we leave to future work.

\paragraph{Future directions: finer-grained steering vectors.}
The directions we use in this work are coarse correctness summaries, namely the GT-vs-Pred displacement in SFT and the correct-vs-wrong contrast in RL. A natural next step is to partition the contrast populations more finely, for example, by execution outcome on code, by reasoning style, by task subtype, or by specific failure mode. Each finer direction could be activated selectively to shape \emph{what} the model learns at a per-skill or per-mistake level rather than uniformly correcting toward a single reference style. The same machinery could also control the \emph{learning dynamics} themselves, varying which directions fire and at what strength as training progresses, rather than fixing a single $\alpha$ and direction throughout. We see steering vectors as a general controller of post-training learning dynamics, of which the two regimes in this paper are only the first instances.

\paragraph{Limitations.}
Our study has several limitations. First, all experiments use student models at or below 7B parameters; whether the findings transfer to larger scales remains open. Second, the best injection layer is model-specific; the fixed mid-stack rule is a reasonable default and the quarter-epoch validation sweep of \cref{app:hparam} finds a better layer cheaply, but there is no principled criterion that predicts it without training. Third, our baseline comparison does not cover on-policy distillation, scheduled sampling, or gold-rollout repair, and in RL, SV-Rescue alone is weaker than entropy-aware shaping alone.

\section{Broader Impact}

Our work studies activation-steering as a training signal for LLM post-training. The positive impact is a potentially better way of post-training. The negative impact is the general dual-use concern that any improvement to LLM reasoning capability can, in principle, be applied to harmful tasks. Our methods do not introduce a new direct path to such misuse beyond what already exists for the public base models we fine-tune. The SV directions we use encode model-vs-reference correctness rather than safety-relevant attributes such as refusal or compliance, so the technique is not a method for bypassing safety alignment.

\end{document}